\documentclass[times,twocolumn,final,authoryear]{elsarticle}
\RequirePackage{geometry}
\long\def\@makecaption#1#2{
	\vskip\abovecaptionskip\footnotesize\bfseries
	\sbox\@tempboxa{#1.~#2}
	\ifdim \wd\@tempboxa >\hsize
	#1.~#2\par
	\else
	\global \@minipagefalse
	\hb@xt@\hsize{\hfil\box\@tempboxa\hfil}
	\fi
	\vskip\belowcaptionskip}

\global\bibsep=0pt

\journal{Computers and Electronics in Agriculture}

\usepackage{graphicx}
\usepackage{pstricks}
\usepackage{multido}
\usepackage[utf8]{inputenc}
\usepackage{amssymb}
\usepackage{amsmath}
\usepackage{latexsym}
\usepackage{subfigure}
\usepackage{hyperref}
\usepackage{setspace}
\usepackage{url}
\usepackage{xcolor}
\usepackage{pdfpages}
\usepackage{lineno}
\usepackage{fancyhdr}
\begin{document}
	
	\title{BioLeaf: a professional mobile application to measure foliar damage caused by insect herbivory}
	
	\author[1,2]{Bruno Brandoli Machado\corref{cor1}}
	\author[1]{Jonatan Patrick Margarido Orue}
	\author[1]{Mauro dos Santos de Arruda}
	\author[1]{Cleidimar Viana dos Santos}
	\author[3]{Diogo Sarath}
	\author[1]{Wesley Nunes Goncalves}
	\author[3]{Gercina Goncalves da Silva}
	\author[3]{Hemerson Pistori}
	\author[3]{Antonia Railda Roel}
	\author[2]{Jose F Rodrigues-Jr}
	
	\cortext[cor1]{Corresponding author: 
		Tel.: +55 67 3437 1730;
		Fax: +55 67 3437 1730;}
	\ead{brunobrandoli@gmail.com}
	
	\address[1]{Federal University of Mato Grosso do Sul, Rua Itibir\'e Vieira, s/n, Ponta Por\~a - MS, 79907-414, Brazil}
	\address[2]{University of Sao Paulo, Av. Trabalhador Sancarlense, 400, Sao Carlos - SP, 13566-590, Brazil}
	\address[3]{Dom Bosco Catholic University, Av. Tamandaré, 6000, Campo Grande - MS, 79117-900, Brazil}
	
	\begin{keyword}
		foliar herbivory, leaf area measurement, plant-herbivore interactions, plant defoliation analysis, feeding injury
	\end{keyword}
	
	\begin{abstract}
		Soybean is one of the ten greatest crops in the world, answering for billion-dollar businesses every year. This crop suffers from insect herbivory that costs millions from producers. Hence, constant monitoring of the crop foliar damage is necessary to guide the application of insecticides. However, current methods to measure foliar damage are expensive and dependent on laboratory facilities, in some cases, depending on complex devices. To cope with these shortcomings, we introduce an image processing methodology to measure the foliar damage in soybean leaves. We developed a non-destructive imaging method based on two techniques, Otsu segmentation and Bezier curves, to estimate the foliar loss in leaves with or without border damage. We instantiate our methodology in a mobile application named BioLeaf, which is freely distributed for smartphone users. We experimented with real-world leaves collected from a soybean crop in Brazil. Our results demonstrated that BioLeaf achieves foliar damage quantification with precision comparable to that of human specialists. With these results, our proposal might assist soybean producers, reducing the time to measure foliar damage, reducing analytical costs, and defining a commodity application that is applicable not only to soy, but also to different crops such as cotton, bean, potato, coffee, and vegetables.
	\end{abstract}
	
	\maketitle
	
	\section{Introduction}
	
	The foliar herbivory quantification is an important source of information on crop production. That is because the foliar area of plants is directly related to the photosynthesis process, which occurs by the incidence of light energy on leaves. As a consequence, the leaves are responsible for the plant growth and grain filling. There are several studies on the effects of foliar herbivory, ranging from forecasting production \citep{straussAN2001,lizasoFCR2003}, molecular and biochemical analysis related to plant defense \citep{kesslerARPB2002,warPSB2012,fescemyerIBMB2013,ankalaPS2013,miresmailliTPS2014}, artificial defoliation analysis \citep{lehtilaES2004,suskoPE2009,johnsonFE2011,liPO2013}, plant fitness in transgenic cultivars \citep{letourneau2009,grinnanEE2013}, and plant invasions on ecological studies \citep{andrieu2011,pirkER2012,moreiraPO2014,croninE2015,calixto2015}. Therefore, quantifying the damage caused by insect herbivory is important with respect to assisting experts and farmers to take better decisions, including evaluations of insecticide management.
	
\begin{figure*}[!ht]
		\centering
		\subfigure[\label{fig:visual} Visual evaluation.]{\includegraphics[width=0.23\textwidth]{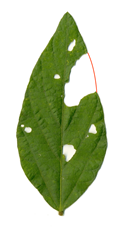}}
		\quad
		\subfigure[\label{fig:manual} Manual quantification.]{\includegraphics[width=0.23\textwidth]{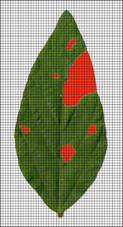}}
		\quad
		\subfigure[\label{fig:limbo} Determination of foliar dimensions.]{\includegraphics[width=0.23\textwidth]{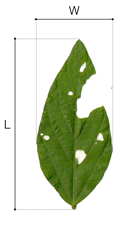}}
		\caption{Different methodologies to estimate foliar damage in a soybean leaf: (a) visual assessment that can be done by an expert; (b) manual quantification using the square-counting method; and, (c) determination of two foliar dimensions.}
		\label{fig:metodos}
	\end{figure*}

	There exist four traditional methodologies to measure foliar damage area: (i) visual evaluation, (ii) manual quantification, (iii) determination of foliar dimensions, and (iv) use of automatic area-integrating meter \citep{licor2014,adc2013}. Assessments are carried out by an expert that, in many cases, estimates wrongly the foliar area (Figure \ref{fig:visual}). The manual quantification, type (ii) is based on the square-counting method \citep{kvet1971}. Typically, an expert or an agronomist counts how many squares fulfill the foliar area over a 1 mm$^2$-spaced grid. This method not only demands an extensive work, but it is time-consuming. Figure \ref{fig:manual} shows an example of how the square-counting method is employed. In turn, method type (iii), determination of foliar dimensions, aims at measuring foliar dimensions based on width and weight measures \citep{cristoforiSH2007,keramatlouSH2015}, as can be seen in Figure \ref{fig:limbo}. Such methodology is widely used by experts, however, it cannot estimate precisely the areas with foliar damages. At last, method (iv) uses devices to automatically measure the leaf area. Although they are accurate to measure leaf areas in case there is no foliage damage, they do not perform well in leaves with insect predation along the borders. Another disadvantage is that they depend on high-cost devices, which demand maintenance. Two examples of such devices are the LI-COR 3100 C \citep{licor2014} and the ADC AM350 \citep{adc2013}.
	
	In order to overcome the drawbacks of the traditional methodologies, there are studies that proposed automatic computational systems. Initial attempts \citep{igathinathaneCEA2006} \citep{easlonAPS2014} are able to measure the foliar area, but do not estimate the damage. Bradshaw {\it et al.} and Neal {\it et al.} \citep{nealJEE2002,bradshawJKES2007} proposed the use of scanners to measure the foliar area. Marcon {\it et al.} \citep{marconRBEAA2011} proposed a similar system for the coffee crop. However, they do not estimate the area with lesions caused by insects, but only the area of healthy leaves so to estimate productivity.

	Mura {\it et al.} \citep{muraSIB2007} proposed an automatic system to estimate the soybean foliar area digitized with the aid of a scanner. Although the automatic system is accurate, it does not estimate the foliar attacked by insects. Furthermore, the system is sensitive to noise, such as sand grains and small pieces of leaves. Similarly, studies of \citep{nealJEE2002,bradshawJKES2007,marconRBEAA2011} are not able to handle noise. In contrast, Nazare {\it et al.}\citep{nazareIWSSIP2010} proposed a methodology that achieves better results even when noise is observed. 
	
	In the work of Easlon and Bloom \citep{easlonAPS2014}, the authors proposed a mobile application, named Easy Leaf Area, which uses image analysis to measure the plant canopy area. That is, different from our application, they do not measure the biomass loss caused by insect herbivory. Not only that, their strategy to segment images is very sensitive to illumination changes. For that reason, in their experiments, they were not able to measure non-green leaves. In nature, leaves have different spots because of the effects of climate and because of insects whose attacks will not necessarily reduce the area of the leaves, rather they just create non-green colors of several shades on the leaves’ surfaces.
	
	In this article, we propose an approach to automatically quantify the foliar damage of insect herbivory by using image processing techniques. In contrast to the other computational proposals aforementioned, we do not use scanners for image acquisition. Here we propose a portable application for smartphones to estimate the damage percentage based on foliar area. Unlike the literature papers, our proposal is able to estimate the area with insect predation, as well as regions in which contours were lost -- we use interactive reconstruction via Bezier curves. With the application, the expert can estimate the damage to the crop {\it in situ}, i.e., there is no need to return to the laboratory. Similar to the proposal of Nazare {\it et al.} \cite{nazareIWSSIP2010}, our application can deal with the noises that appear in the images, eliminating them using connected components \cite{gonzalez2006}. Although Nazare has proposed a methodology that makes the reconstruction of contours, it is based on line segments, so that the reconstruction depends on the corners that the algorithm detects. Hence, in Nazare's method, instead of a curve, the contour of attacked regions is filled with a set of line segments, turning the measurement inaccurate. In the work of Bradshaw {\it et al.} \citep{bradshawJKES2007}, they estimate the attacked regions, but, since the authors used polygons to estimate the area that suffered the insect herbivory, their results are not accurate. Differently, here we use quadratic Bezier curves with three user-defined control points, creating a smooth contour that fits the original edge of leaves. Furthermore, our application can be used as a non-destructive method because it does not require leaf removal from the plant, which allows repeated measurements of the same leaf.
	
	\section{BioLeaf - Foliar Analysis: A Novel Approach to Estimate Leaf Area Loss}
	\label{sec:approach}
	
\begin{figure*}[!ht]
		\centering
		\subfigure[\label{fig:propostaA}]{\includegraphics[width=0.24\textwidth]{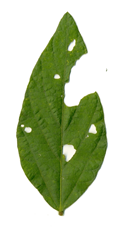}}
		\quad
		\subfigure[\label{fig:propostaB}]{\includegraphics[width=0.24\textwidth]{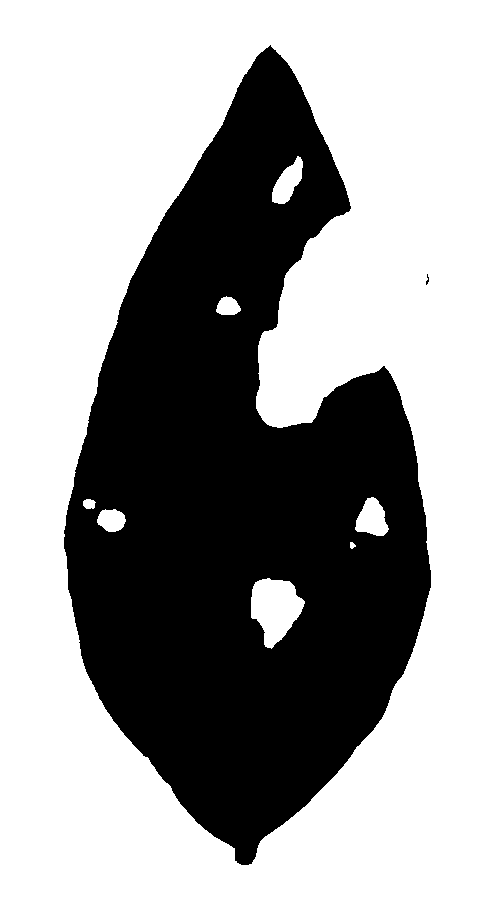}}
		\quad
		\subfigure[\label{fig:propostaC}]{\includegraphics[width=0.24\textwidth]{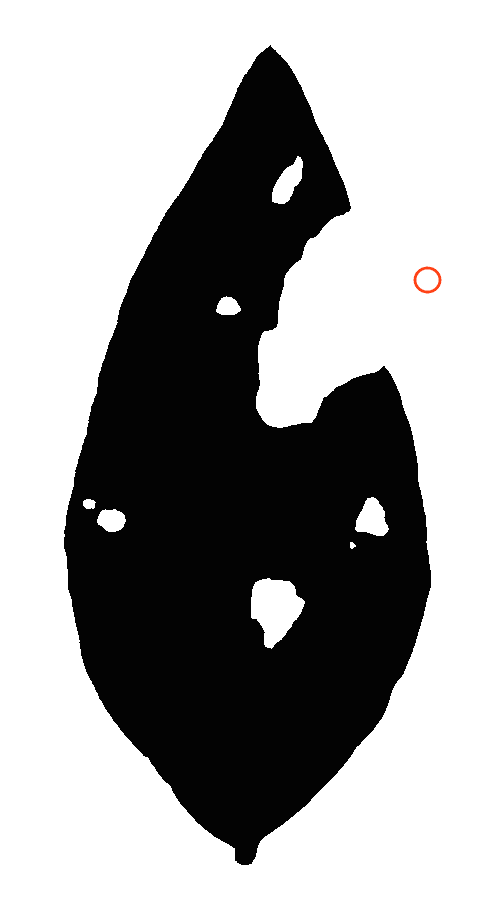}}
		\\
		\subfigure[\label{fig:propostaD}]{\includegraphics[width=0.24\textwidth]{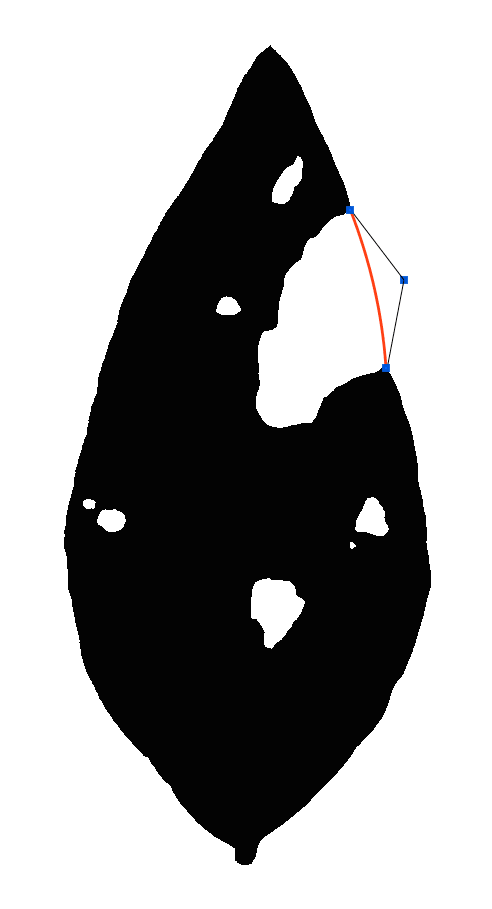}}
		\quad
		\subfigure[\label{fig:propostaE}]{\includegraphics[width=0.24\textwidth]{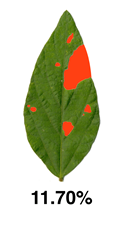}}
		\caption{\label{fig:proposta} The proposed approach: (a) original leaf; (b) segmented leaf; (c) noise removal; (d) three control points and Bezier line; and (e) quantification of damaged foliar area.}
	\end{figure*}

	In this section, we introduce a methodology to estimate the foliar damage of leaves. Our goal is to calculate the intensity of foliar loss in relation to the total leaf area. Figure \ref{fig:proposta} illustrates the method that consists of four steps based on techniques of image processing and computational geometry applied to each leaf image: (i) image thresholding, (ii) noise removal,  (iii) border reconstruction using quadratic Bezier curves and, (iv) insect herbivory quantification. 
	
	After the image acquisition step (Figure \ref{fig:propostaA}), we perform the image processing using image thresholding. Basically, thresholding is an image segmentation technique that considers the criterion of image partitioning. This criterion defines one or more values, called thresholds, that divide the image in objects of interest and background. This technique is used in many works \citep{rosinPRL2003,sankurJEI2004} and applications \citep{mizushimaCEA2013,kurtCMPB2014,wangJNM2015}. Typically, there are two groups of thresholding methods. Global thresholding uses one single threshold for the entire image, while local thresholding uses different thresholds depending on the region that will be segmented. In this work, we used the global thresholding of Otsu's method \citep{otsuIEEETSMC1979}, which has already been used to segment leaves with good performance. We show that the same principle of Otsu can be applied in mobile devices. Figure \ref{fig:propostaB} shows the image segmentation result. However, inadequate illumination is a major problem in image processing because uneven illumination has impact in the result of segmentation operations. To tackle this problem in our application, we used the Otsu's method on the CIE La*b* color space \citep{gonzalez2006}. We convert an image from RGB to CIE La*b* and use it to reduce the impact in cases when unbalanced light reflection occurs on the leaf's surface. Furthermore, the resulting image is shown on the screen, and if the user realizes it is not satisfactorily segmented, he still can fine-tune the automatic segmentation.
	
	Despite the efficacy of thresholding for image processing, there might be leaves with small pieces of leaves and sand grains, or with damaged borders caused by insect herbivory. In such cases, segmentation is not enough. Hence, we use connected components to automatically remove noise that can appear after the image was captured. Connected components labeling works by grouping pixels of an image based on the adjacency of pixels that share the same intensity \citep{gonzalez2006}. In this paper, we assume binary input images and 8-connectivity. Once groups are formed, they are filtered by their size. Figure \ref{fig:propostaC} illustrates this step of how the application removed automatically a small group of pixels (circle highlighted in red color).
	
	In the proposed application, we also use a computational geometry technique, named parametric Bezier curves, to model and reconstruct the borders of damaged leaves. The interactive reconstruction is necessary because many times insects can damage the borders of the leaves, as it can be seen in Figure \ref{fig:propostaD}. In this step, the expert is supposed to create the control points by touching the screen (squares highlighted in blue color) in order to adjust the curve to the leaf's curve. It is worth noting that, in some cases, the insects can attack more than one border and the expert should create more than one curve. Finally, in Figure \ref{fig:propostaE}, we calculate the percentage of herbivory by counting the pixels which stand for insect predation in relation to the total leaf area.
	
	\section{Materials and Methods}
	\label{sec:meto}
	
	\subsection{Experimental Design}
	\label{sec:exps}
	
	In the experiments, we collected leaves of soybean (\textit{Glycine max (L.) Merrill}) from natural and transgenic plants. We considered both groups because, through breeding, soybean can be genetically engineered, therefore, observing their resistance to herbivory is an interesting analytical task. We divided the collected leaves into three groups: (1) leaves with existent attacks from caterpillars; (2) leaves with artificial defoliation, i.e., defoliation-mimicking herbivory to simulate insect attack; and (3) control group, that is, leaves collected with no injury in the greenhouse, and then, exposed for the first time to caterpillars. Then, each group was separated into transgenic and non-transgenic, resulting in six groups of leaves.
	
	In order to build the groups damaged by caterpillars, we collected caterpillars of species \textit{Spodoptera frugiperda}) from a soybean crop. Such caterpillars were left fasting for 24 hours, after that they were exposed to group 1 leaves, both transgenic and non-transgenic for another 24 hours. Lastly, we captured images of six leaves from each group, three images per leaf, resulting in 36 images. We used a Sony Alpha DSLR-A350 14.2 MP camera without flash, positioned 30 cm from the leaves, and having a white portable background to be used in the crop, preventing the removal of leaves. The images were captured with $1024 \times 1024$ pixels and stored according to the TIFF format.
	
	\subsection{Image segmentation by Otsu's Method}
	
	The method of Otsu is a cluster-based image segmentation that converts a gray-level image into a binary image. That is, it assumes that the image contains two classes of pixels, calculating the optimum threshold $T$ that separates the two classes so that their intra-class variance is minimal (or, that their inter-class variance is maximal) \citep{otsuIEEETSMC1979}. The method computes the probability density function of the gray-level image, assuming bi-modal Gaussian distributions represented as discrete histograms. The histograms are represented as uni-dimensional vectors whose bins refer to the intensity levels of the pixels; as so, given the intensity $i$ of a pixel in a $M \times N$ image, the probability density function $P_i$ is given by: 
	
	\begin{equation}
	P_{i} = \frac {n_{i}}{M \times N}
	\label{eq:p}
	\end{equation}
	
	\noindent where $n_i$ is the number of pixels with intensity $i$, $0 \leq i < L$, and $L$ is the maximum of gray-level.
	
	Formally, the technique assumes a threshold $T$, $0 \leq i < L$, which separates the pixels into two classes according to their intensities. Class $C_0$ with pixels whose intensities vary in the range $[0,T-1]$, and class $C_1$ with pixels in the range $[T,L-1]$. The method computes $T$ by varying its value and calculating the inter-class variance $\sigma$ for each value according to Equation \ref{eq:otsu-equation}. The highest variance indicates the optimal value of $k$:
	
	\begin{equation}
	\label{eq:otsu-equation}
	\sigma^2_{k} = \omega_{0}(\mu_{0} - \mu_{T})^2 + \omega_{1}(\mu_{1} - \mu_{T})^2
	\end{equation}
	
	\noindent where 0 and 1 are indexes referring to classes $C_0$ and $C_1$, and $\mu_{0}$, $\mu_{1}$, $\omega_{0}$, and $\omega_{1}$ are given by the equations that follow:
	
	\begin{equation}
	\omega_{0} = \sum_{i=0}^{k-1} P(i)
	\quad\mbox{and}\quad
	\omega_{1} = \sum_{i=k}^{L-1} P(i)
	\end{equation}
	
	\begin{equation}
	\mu_{0} = \sum_{i=0}^{k-1} i P(i)/\omega_{0}
	\quad\mbox{and}\quad
	\mu_{1} = \sum_{i=k}^{L-1} i P(i)/\omega_{1}
	\end{equation}
	
	Finally, the global mean $\mu_{T}$ is given by:
	
	\begin{equation}
	\mu_{T} = \sum_{i=0}^{L-1} i P(i)
	\end{equation}
	
	The result of the segmentation is a binary image with black pixels for the object of interest; in our context, the leaf becomes black over a white background. The image segmentation is necessary for our process; however, note that, in our application, we draw the leaf with its original color.
	
	\subsection{Bezier-based leaf reconstruction}
	
	Segmentation provides a manner to identify damage in leaves. However, if the border of the leaf is damaged, segmentation fails as it is not able to distinguish background from leaf. To solve this problem, we use Bezier curves to interpolate the best curve able to restore the original leaf border. These curves refer to a polynomial function that, based on control points (given by the user) \citep{jordanCAGD2014}, is able to fulfill gaps around a given leaf. The curve can be represented as a binomium of Newton, considering that solving the curve corresponds to setting the coefficients of the binomium for each point $t$ of the curve \citep{fitterPE2014}. Accordingly, the corresponding $B(t)$ point for a point $t \in [0,1]$, following Bezier curve, is given by:
	
	\begin{equation}
	B(t) = \sum\limits_{i=0}^{n} {n\choose i} (1 - t)^{(n - i)}t^i * B_i
	\label{eq:bezier}
	\end{equation}
	
	\noindent where each given $t$ represents the value of parameterization to go through the curve and $n$ is the degree of the binomium -- the technique demands $n+1$ control points. The control points are represented as $B_i$ and ${n\choose i}$ refer to the binomial coefficients.
	
	In our problem, we use three control points, which leads to the quadratic function:
	
	\begin{equation}
	B(t) =  (1-t)^2B_0 + 2t(1-t)B_1 + t^2B_2
	\label{eq:bezierquad}
	\end{equation}
	where the first and last control points are the end points of the curve. In our application, $B_i's$ are set by the user's touch on the borders of the leaf.
	
	The quadratic version demonstrated to be effective in reconstructing the borders of the leaves with reduced computational cost, especially if compared to the cubic version.
	
	\subsection{Application on Mobile Devices}
	
	The use of mobile devices to perfom automatic tasks has increased fast \citep{fengMS2015}. The main reasons for it are the recent advances in hardware, such as sensors, processors, memories, and cameras. Thereby, smartphones have become new platforms for applications of image processing and computer vision \citep{casanovaESA2013,farinellaPR2015}. Furthermore, mobile devices can perform tasks in real-time {\it in situ} far from the laboratory.
	
\begin{figure*}[!ht]
		\centering
		\includegraphics[width=0.9\textwidth]{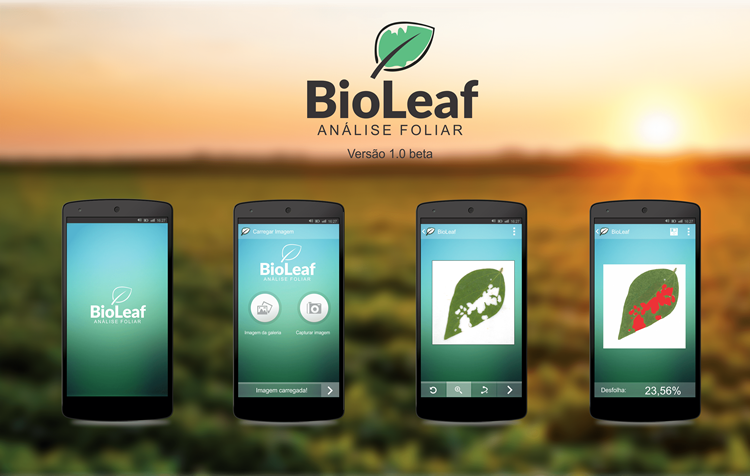}
		\caption{\label{fig:telas} Screenshots of the BioLeaf application, freely available at \url{https://play.google.com/store/apps/details?id=upvision.bioleaf}}
	\end{figure*}

	In this context, besides its methodology, the contribution of this paper is the development of a mobile application to perform automatic quantification of leaf damages caused by insect herbivory. Some screens of the application proposed here, named as BioLeaf - Foliar Analysis, are shown in Figure \ref{fig:telas}. The application was developed in the Java programming language over the Integrated Development Environment Android Studio, following the algorithms described in Section \ref{sec:meto} for platform Android. The application can be downloaded freely from GooglePlay at \url{https://play.google.com/store/apps/details?id=upvision.bioleaf}.
	
	\section{Experimental Results}
	\label{sec:res}
	
	In this section, we evaluate the proposed quantification of foliar damage in comparison to the manual estimation. First, we describe the results of the internal quantification -- when borders are preserved after damage; then, we describe the results when the borders need to be reconstructed with Bezier curves. Experiments were conducted on different groups of leaves, as described in Section \ref{sec:meto}. We used six groups of leaves divided by three types of damages (see Sec. \ref{sec:exps}) and two types of plant breeding (transgenic and non-transgenic), with six images per groups and a total of 36 images.
	
\begin{figure*}[!ht]
		\centering
		\subfigure[\label{fig:internaSpo}\textit{Spodoptera frugiperda} and internal attack only]{\includegraphics[width=0.45\textwidth]{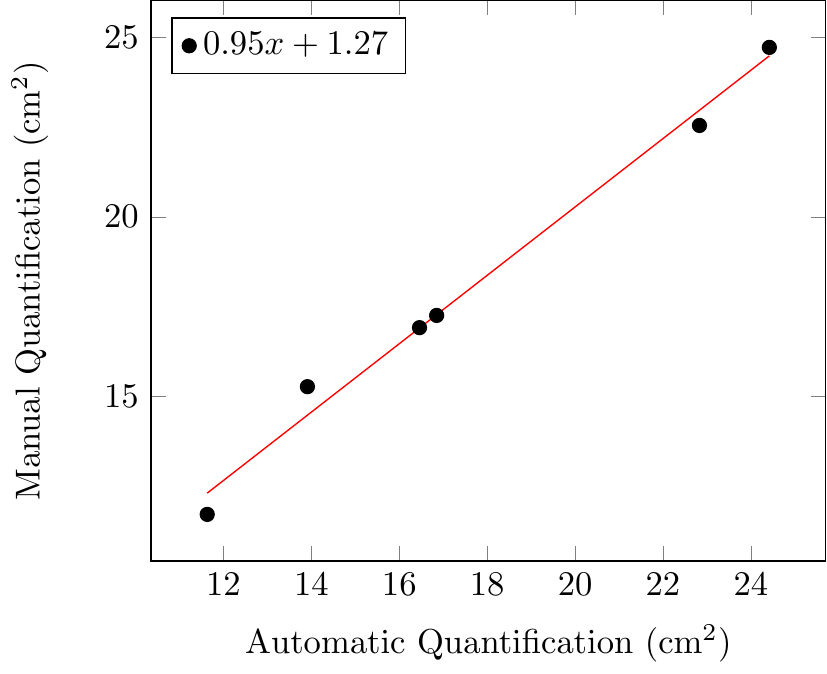}}
		\subfigure[\label{fig:curvaSpo}\textit{Spodoptera frugiperda} and damaged border]{\includegraphics[width=0.45\textwidth]{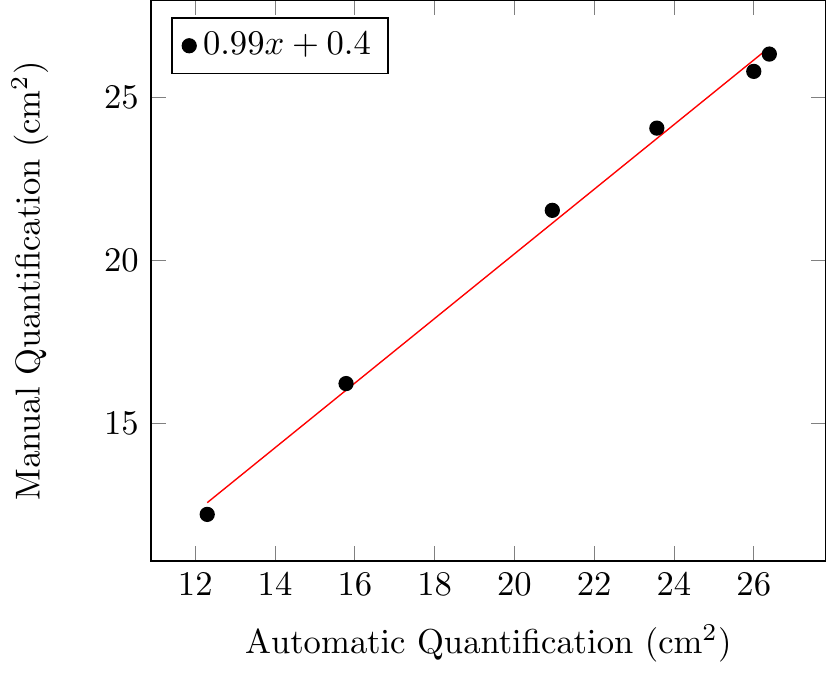}}
		\subfigure[\label{fig:internaManual}Artificial defoliation and internal attack only]{\includegraphics[width=0.45\textwidth]{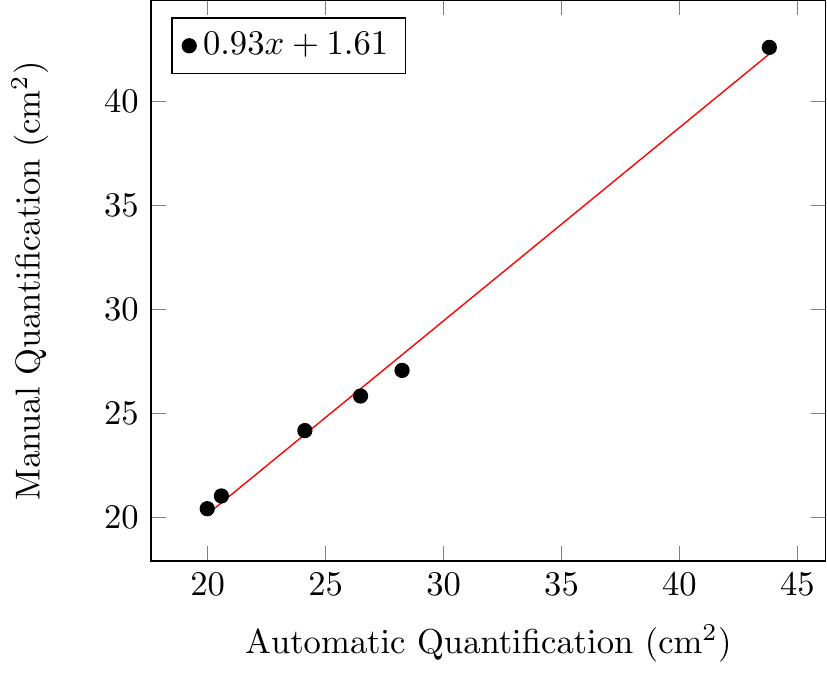}}
		\subfigure[\label{fig:curvaManual}Artificial defoliation and damaged border]{\includegraphics[width=0.45\textwidth]{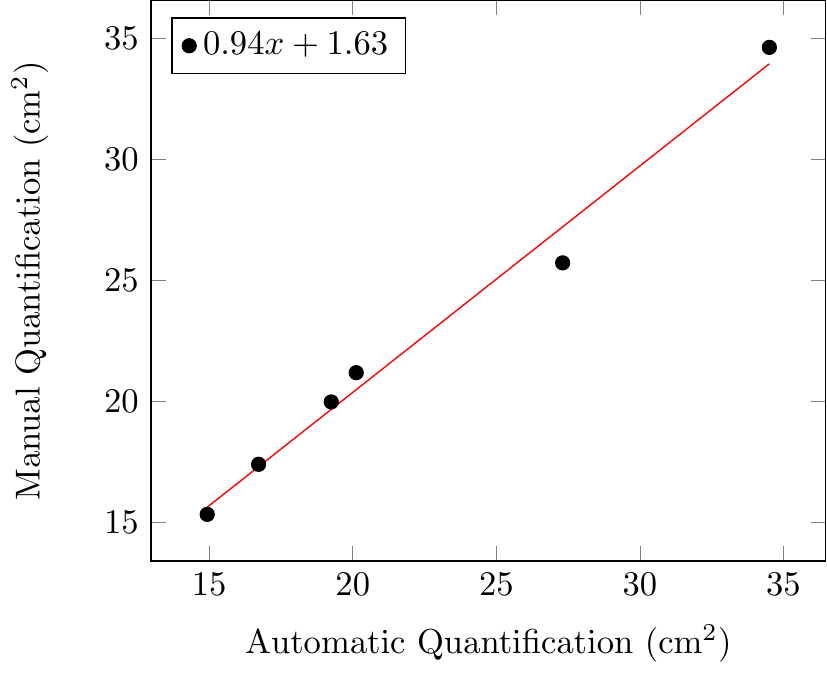}}
		\subfigure[\label{fig:internaControl}Control group and internal attack only]{\includegraphics[width=0.45\textwidth]{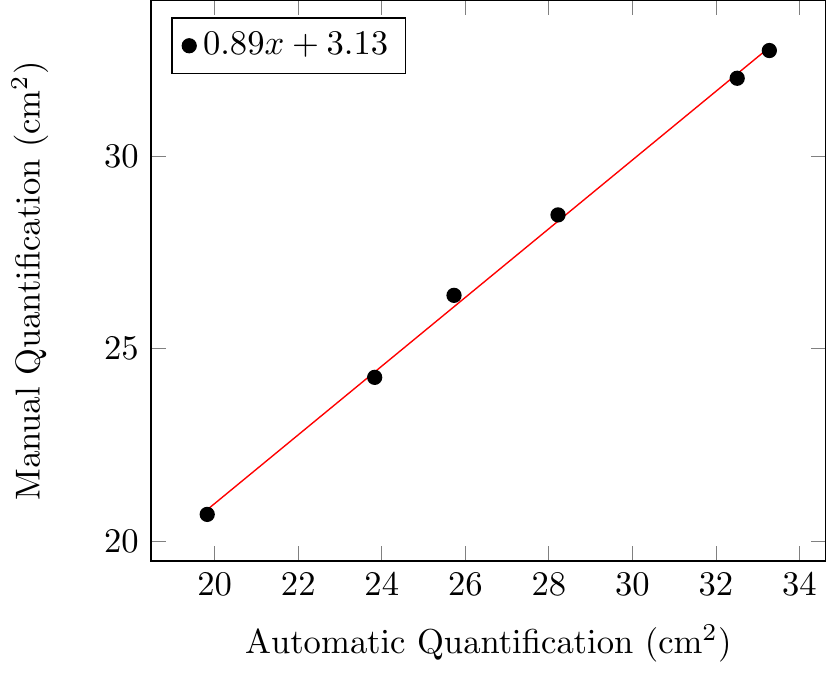}}
		\subfigure[\label{fig:curvaControl}Control group and damaged border]{\includegraphics[width=0.45\textwidth]{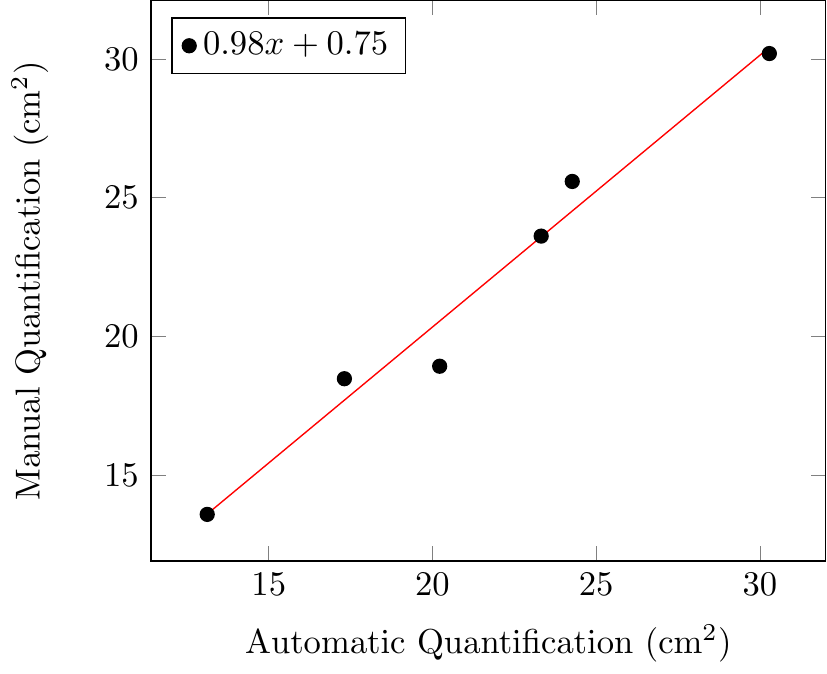}}
		\caption{\label{fig:allplots}Relationship between manual and automatic damage area quantification. Experiment 1 is presented in the left column, (a), (c), and (e), for leaves with internal attack only -- respectively, caterpillar damage, synthetic damage, and control group. Experiment 2 is presented in the right column, (b), (d), and (f), for leaves with damaged borders  -- respectively, caterpillar damage, synthetic damage, and control group.}
	\end{figure*}

	\medskip\noindent\textbf{Experiment 1 - Internal damages:} 
	First, the \textit{internal damage} quantification of the damaged foliar area corresponds to the estimation of the herbivory that affects only the inner parts of the leaves. In these terms, leaf lesion was defined as the ratio between the number of pixels in the damaged areas, and the number of pixels of the entire leaf. Here, we manually and automatically estimated the damaged areas (in cm$^2$) for 18 leaves with internal damage. To evaluate the accuracy, we analyzed the linear correlation \cite{gibbons1985} for the automatic and manual quantifications (Figures \ref{fig:internaSpo}, \ref{fig:internaManual} and \ref{fig:internaControl}) for leaves ranging in size from 11.63 cm$^2$ to 43.81 cm$^2$. 
	No significant divergence was observed at any leaf size. The larger variation was observed in the BioLeaf estimates, in comparison to the square-counting method. It was of only 1.24 cm$^2$  (or 7.30\% of the damage) in the group with artificial defoliation and internal attacks (Figure \ref{fig:internaManual}), as verified in the group of non-transgenic soybean (left column of the figure). The concordance correlation coefficients across all groups were greater than $R \geq 99.5$ with a $P$-$value < 0.001$.
	
	\medskip\noindent\textbf{Experiment 2 - Border reconstruction:} In the second experiment, we quantified the damaged area in leaves with \textit{damaged borders}. The reconstruction was carried out by a specialist who was in charge of setting three control points per border segment. Figure \ref{fig:propostaD} illustrates the process. Similar to the first experiment, in Figures \ref{fig:curvaSpo}, \ref{fig:curvaManual} and \ref{fig:curvaControl}, we see the same plot for leaves with damaged borders. Again, the line is close to the linear model.
	Image sizes ranged from 12.22 cm$^2$ to 34.62 cm$^2$. Effectively, the most significant difference of automatic to manual quantification was of only 1.60 cm$^2$ for the artificial defoliation (Figure \ref{fig:curvaManual}). The foliar damage estimation by BioLeaf was highly similar with significant linear correlation coefficients (at least greater than $R \geq 98.0$ and $P$-$value < 0.001$) correlation coefficients were slightly lower when compared to internal damage only.
	
\begin{figure*}[!ht]
		\centering
		\subfigure[\label{fig:interna}Internal Damage]{\includegraphics[width=0.45\textwidth]{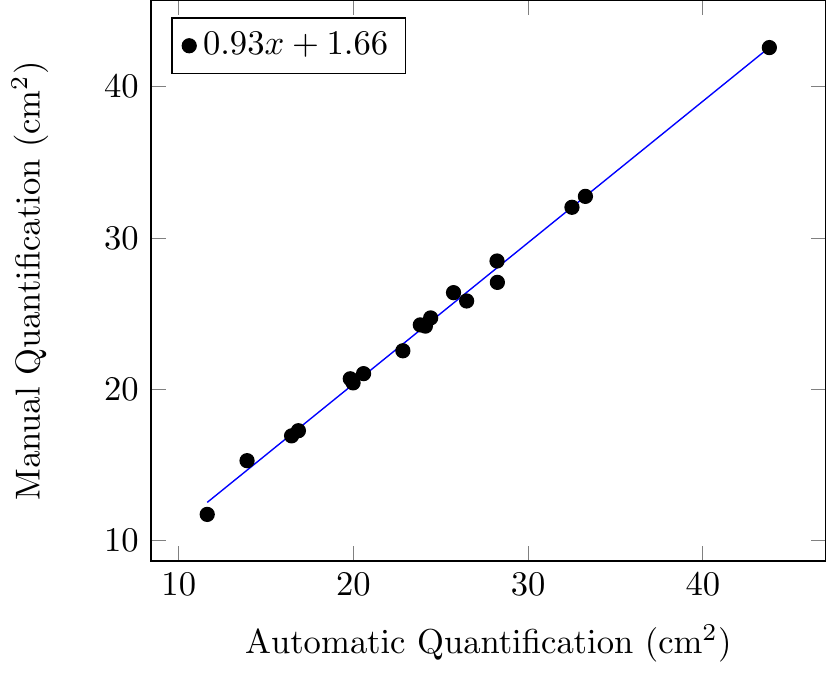}}
		\subfigure[\label{fig:curva} Reconstructed borders]{\includegraphics[width=0.45\textwidth]{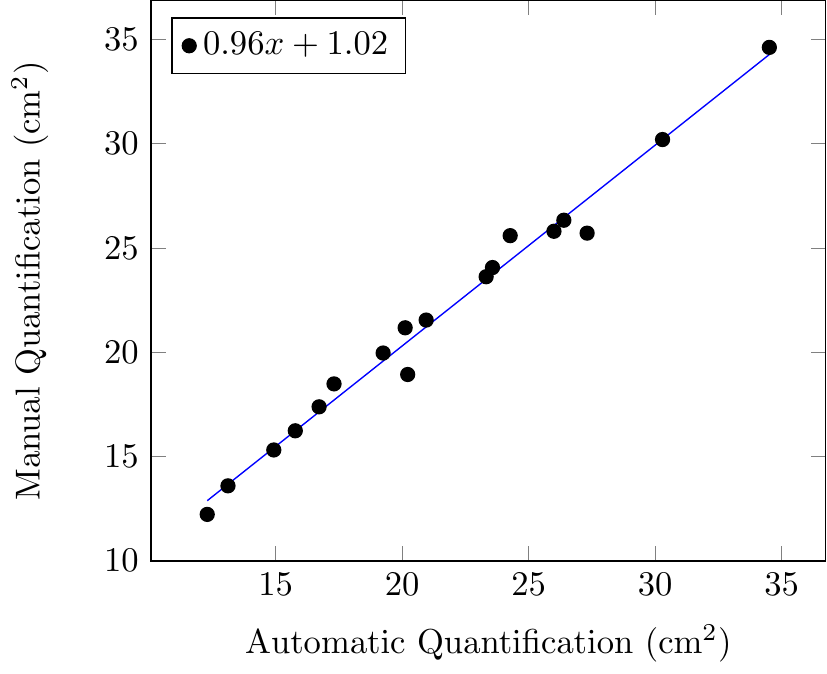}}
		\caption{Linear correlation plot demonstrating high accuracy for (a) leaves with internal damage only; and (b) with damaged borders.}
	\end{figure*}

	\medskip\noindent\textbf{Overall evaluation}: Next, we show the linear correlation plots for all the groups of leaves with internal damage and border damage. Linear correlations were highly similar ($P$-$value < 0.001$) in leaves with borders preserved ($R \geq 99.76$; Figure \ref{fig:interna}), while leaves with reconstructed borders had a slightly smaller correlation, with coefficient of $R \geq 99.24$ and $P$-$value < 0.001$ (Figure \ref{fig:curva}).
	
	\medskip\noindent\textbf{Processing speed}: The time required to measure each leaf, as well as to calculate the damaged area caused by insect herbivory, was substantially less than the manual quantification using the square-counting method. The average image processing time was less than 1 second for internal quantification. In comparison to the square-counting method, which takes between 20 to 30 minutes for each leaf, BioLeaf was nearly 1,500 times faster. It is important to point that we did not consider the preparation time for both automatic and manual once they are the same, i.e., aiming at a fair comparison we just calculate the time of the measurement of the attacked area by insects for both methods, manual and automatic. The average time to process images when the borders needed to be reconstructed was between 10 and 25 seconds. This times tends to reduce as the user becomes more experienced with the application.
	
\begin{figure*}[!ht]
		\centering
		\subfigure{\includegraphics[width=0.2\textwidth]{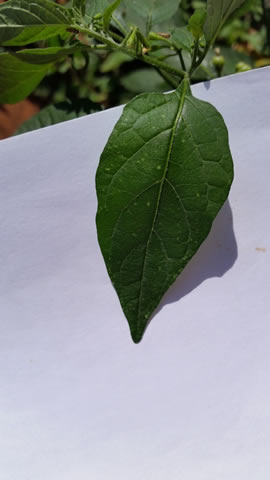}}
		\subfigure{\includegraphics[width=0.2\textwidth]{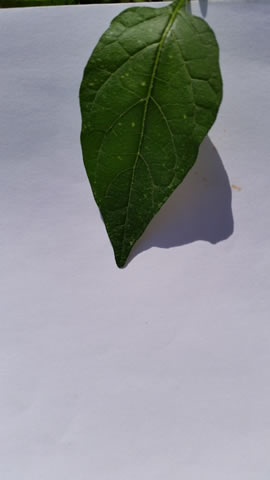}}
		\subfigure{\includegraphics[width=0.2\textwidth]{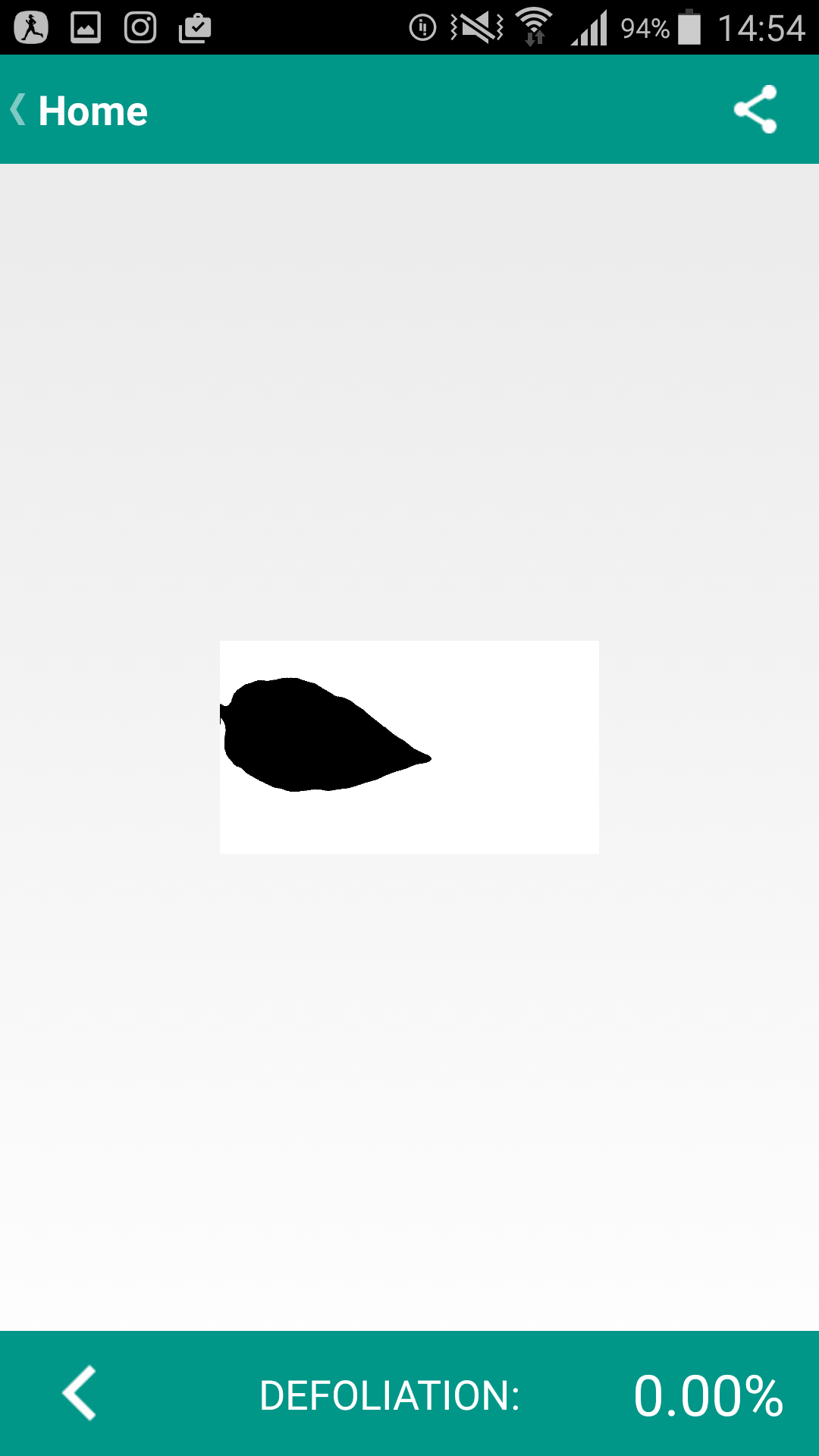}}
		\\
		\subfigure{\includegraphics[width=0.2\textwidth]{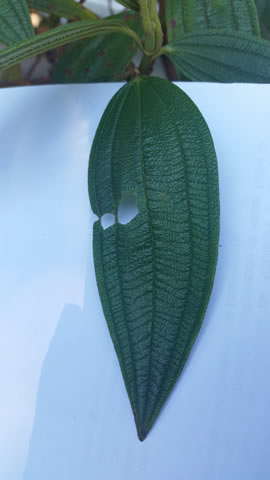}}
		\subfigure{\includegraphics[width=0.2\textwidth]{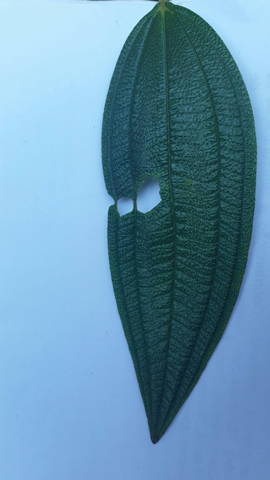}}
		\subfigure{\includegraphics[width=0.2\textwidth]{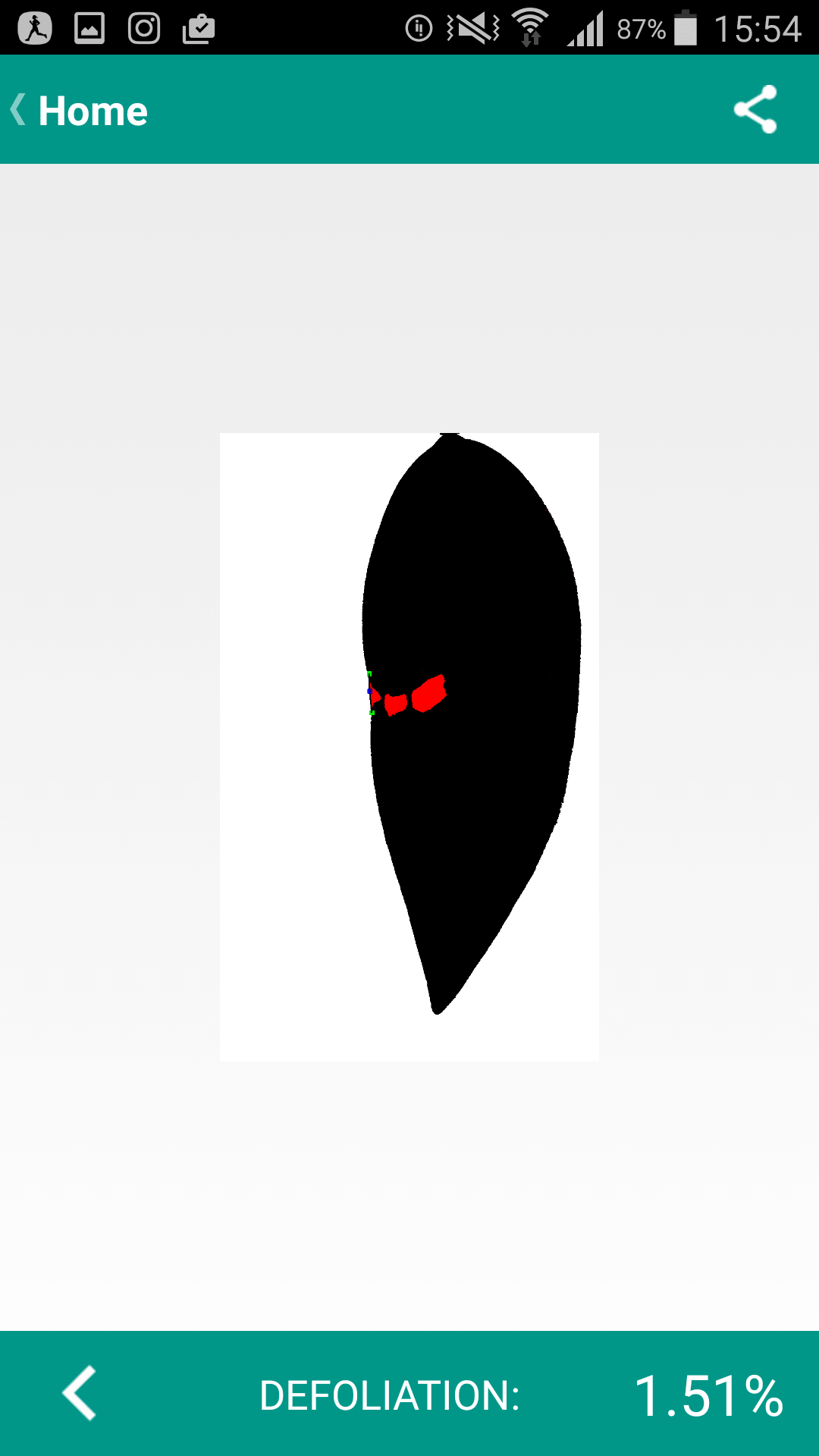}}
		\\
		\setcounter{subfigure}{0}
		\subfigure[No Leaf Removal]{\includegraphics[width=0.21\textwidth]{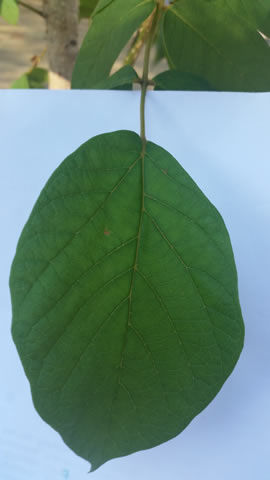}}
		\subfigure[Framing the Leaf]{\includegraphics[width=0.21\textwidth]{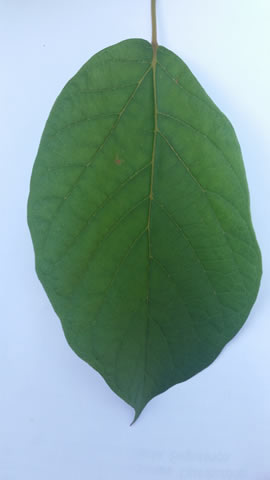}}
		\subfigure[Result with BioLeaf]{\includegraphics[width=0.21\textwidth]{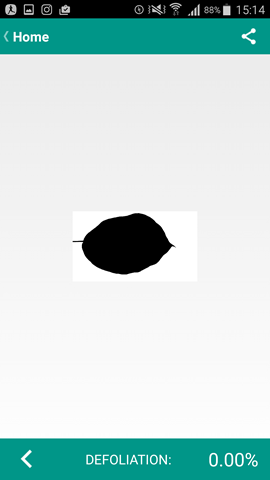}}
		\caption{\label{fig:nonDestructive}Experiments for non-destructive measuruments of three different plants collected from the Brazilian Cerrado.}
	\end{figure*}

	\medskip\noindent\textbf{Non-destructive measurement with BioLeaf}: In this experiment, we show the steps on how our mobile application works for measurements with no leaf removal. To this end, we recommend the expert to use a background with contrast to the leaf when taking the picture. Figure \ref{fig:nonDestructive} illustrates the process for three samples taken under different illumination conditions and different species of plants of the Brazilian Cerrado forestry. Each row shows one species of plant and its respective defoliation estimative. In two cases the defoliation was 0\% since the leaves have no damage.
	
\begin{figure*}[!ht]
			\centering
			\subfigure[Brachiara brizantha, 0.2 meters]{\includegraphics[width=0.2\textwidth]{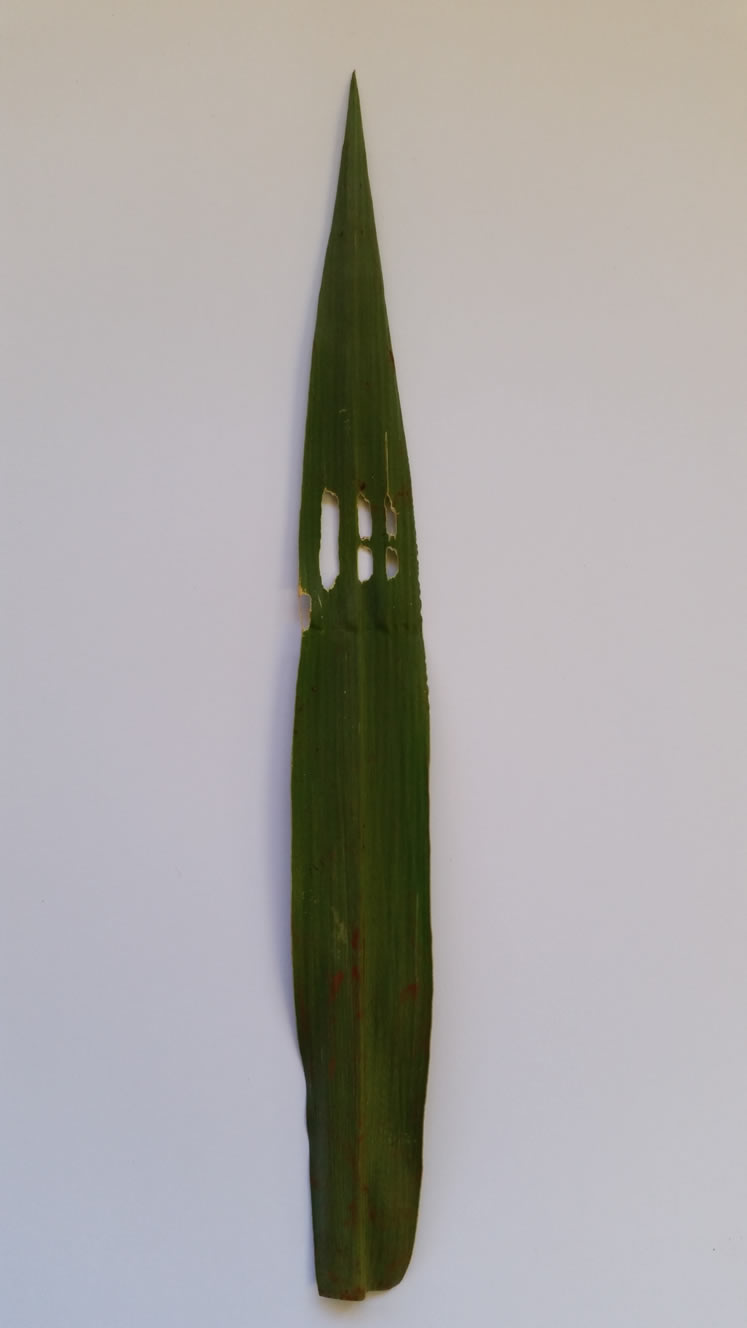}
		    \includegraphics[width=0.2\textwidth]{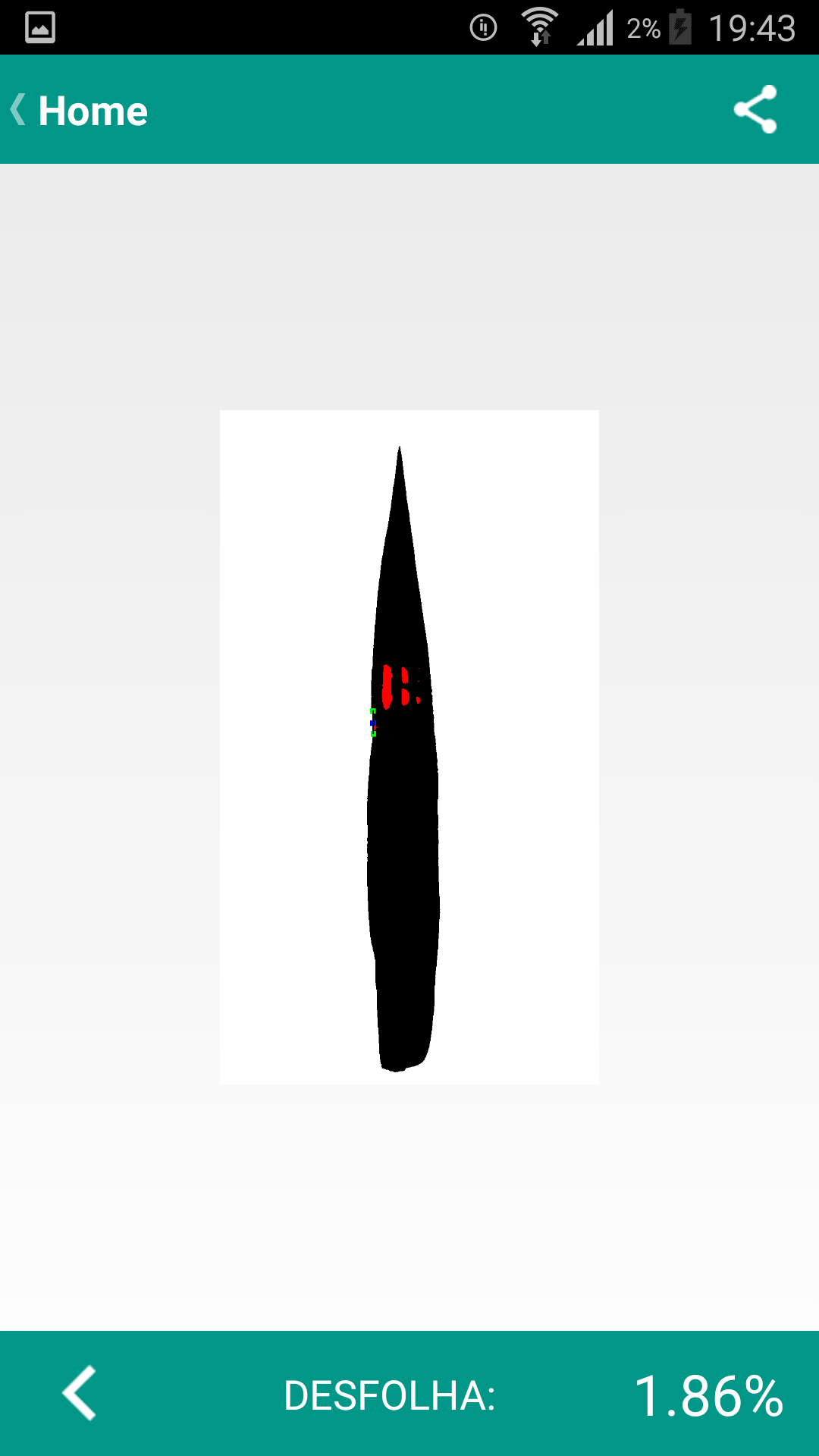}}
			\quad
			\subfigure[Sugar cane, 1.8 meters]{\includegraphics[width=0.12\textwidth]{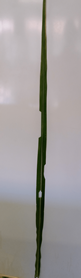}
			\includegraphics[width=0.2\textwidth]{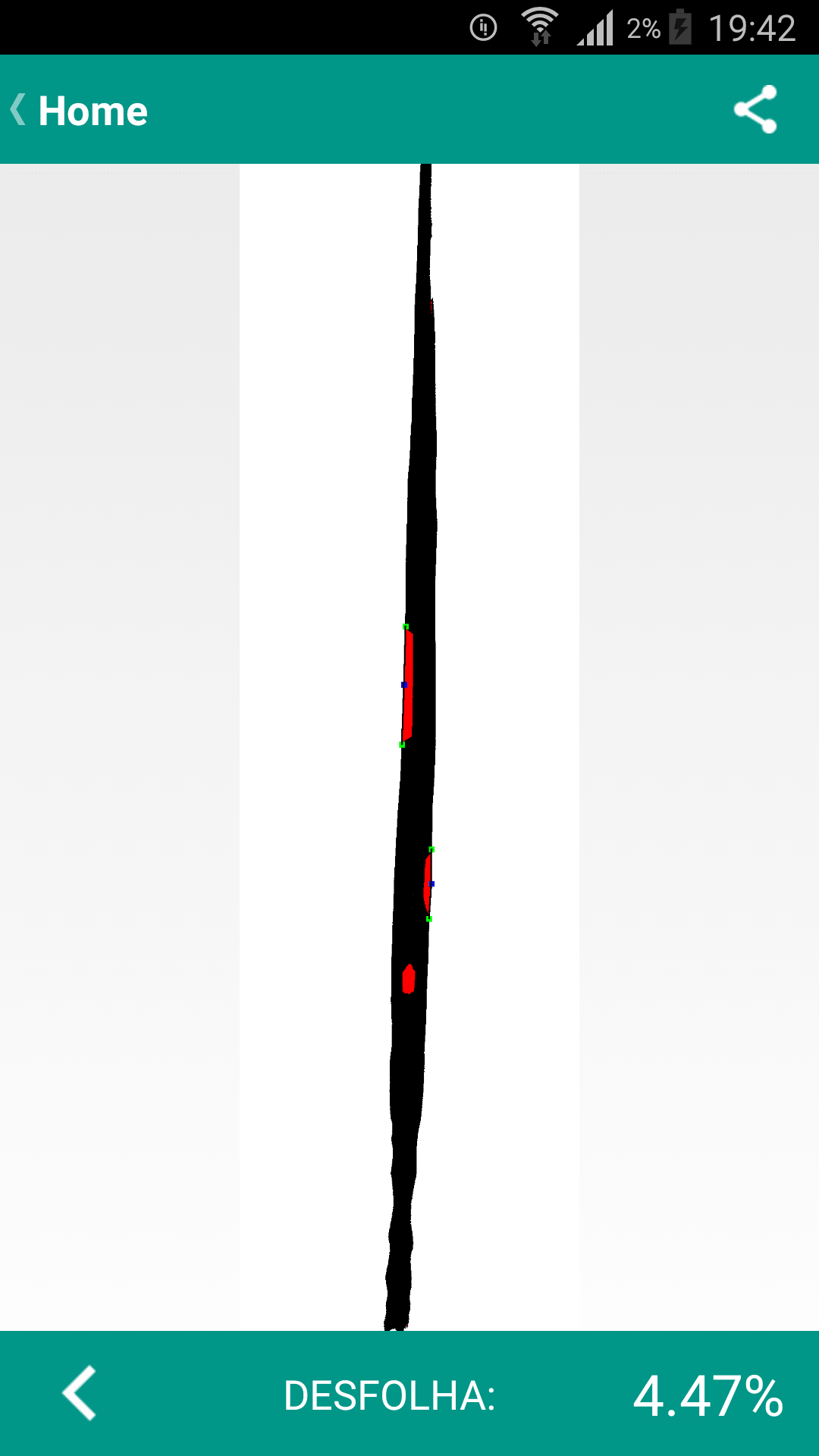}}
			\\
			\subfigure[Brachiara marandu, 0.3 meters]{\includegraphics[width=0.23\textwidth]{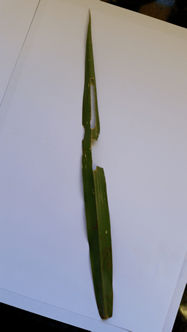}
			\includegraphics[width=0.2\textwidth]{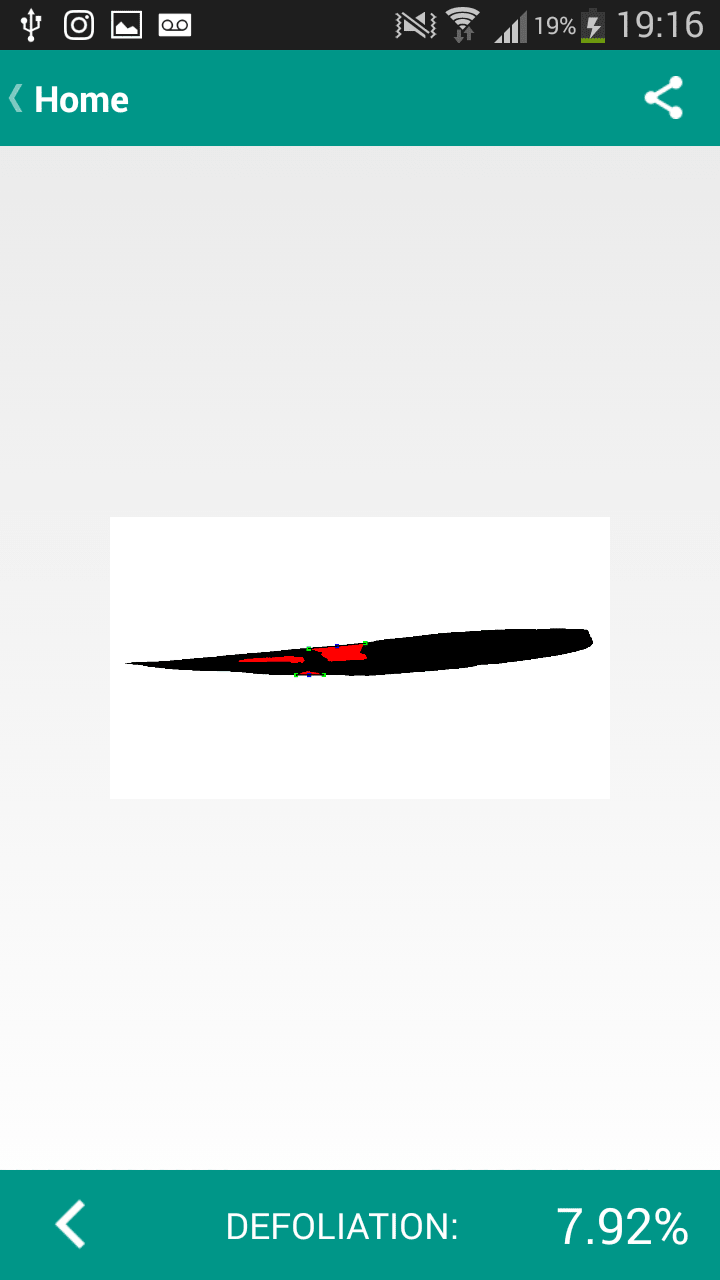}}
		    \quad
			\subfigure[Brachiara marandu, 0.3 meters]{\includegraphics[width=0.23\textwidth]{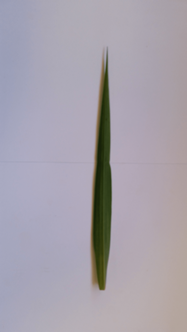}
				\includegraphics[width=0.2\textwidth]{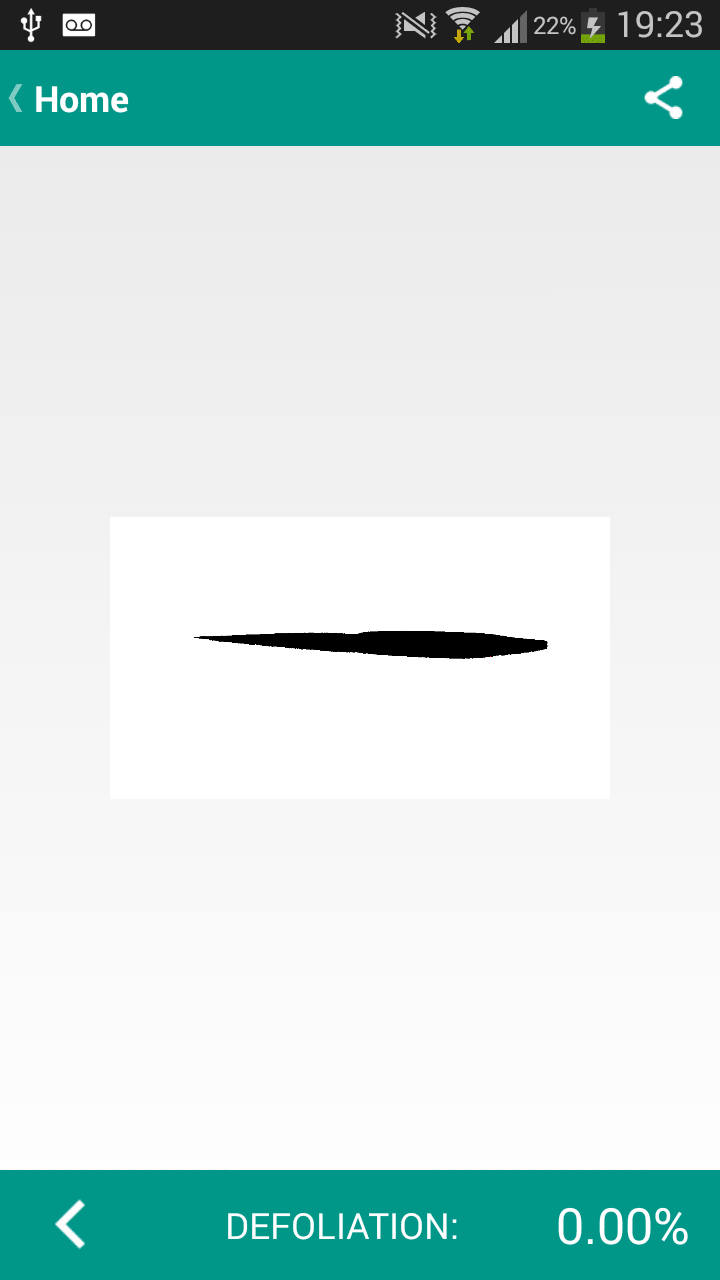}}
			\\
			\subfigure[Panicum maximum, 0.5 meters]{\includegraphics[width=0.2\textwidth]{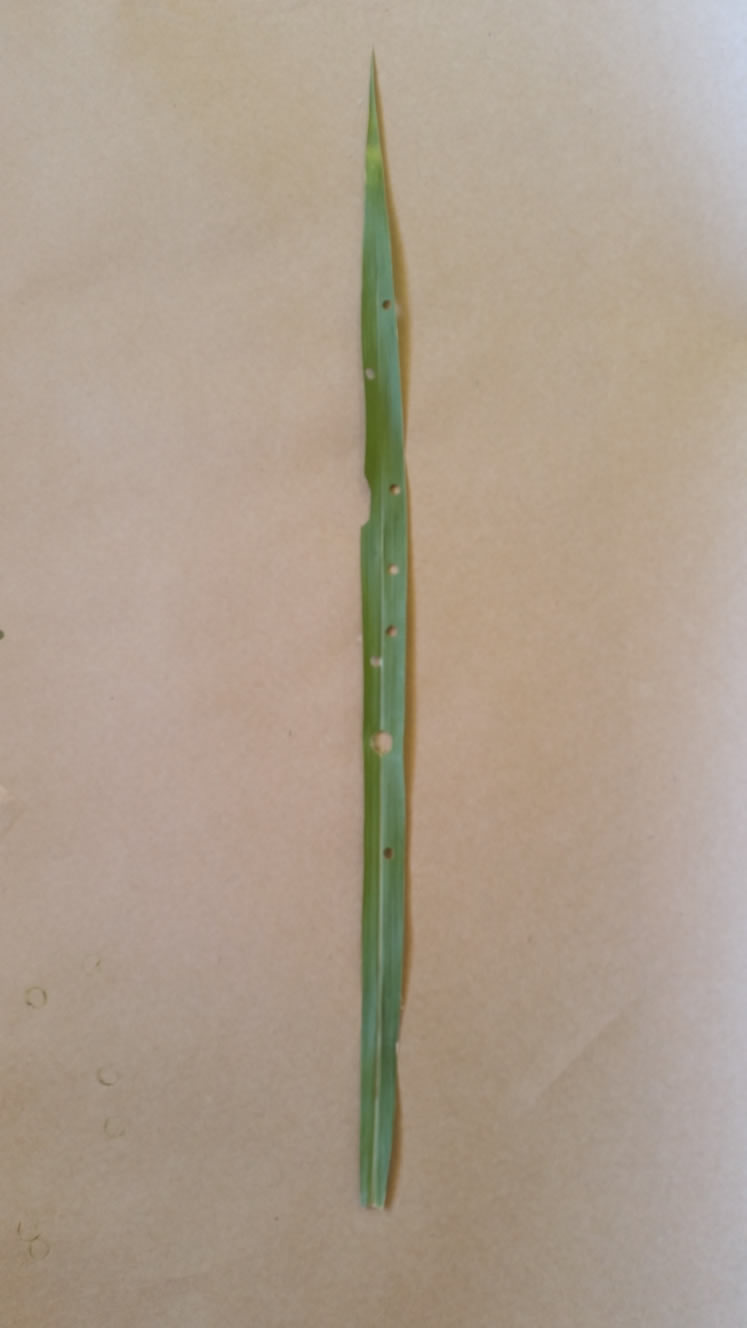}
			\includegraphics[width=0.2\textwidth]{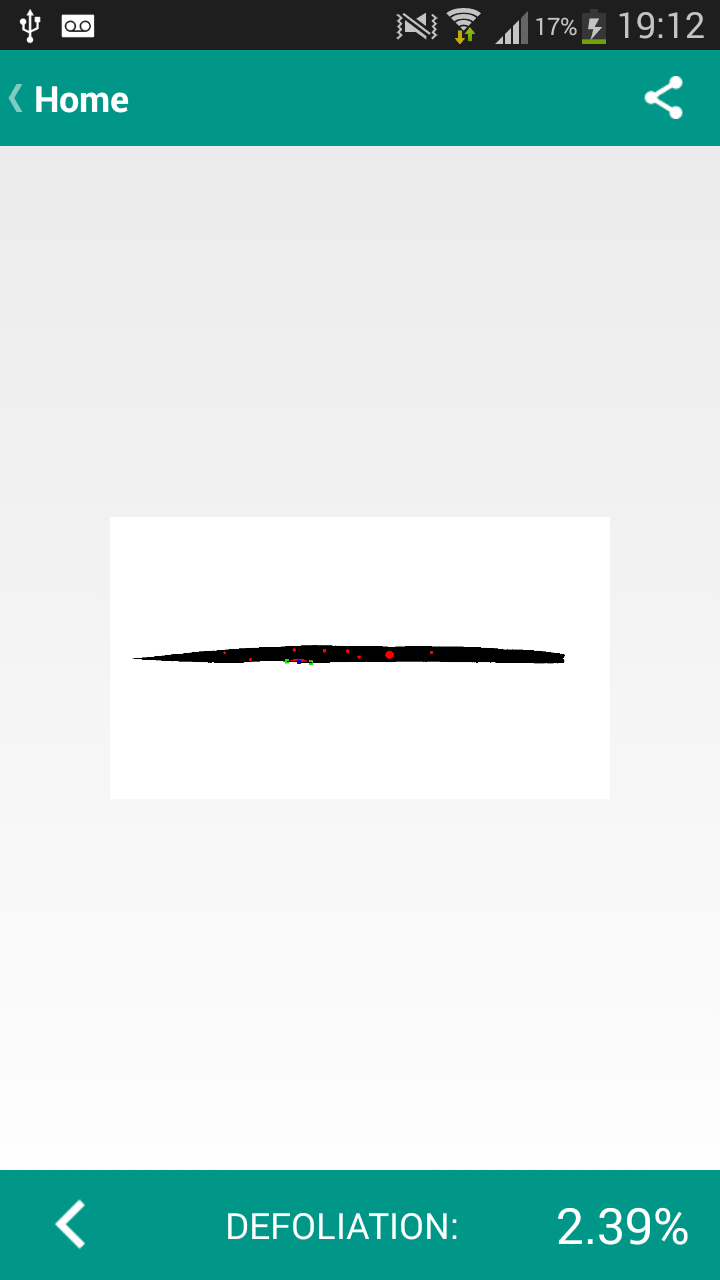}
			\includegraphics[width=0.2\textwidth]{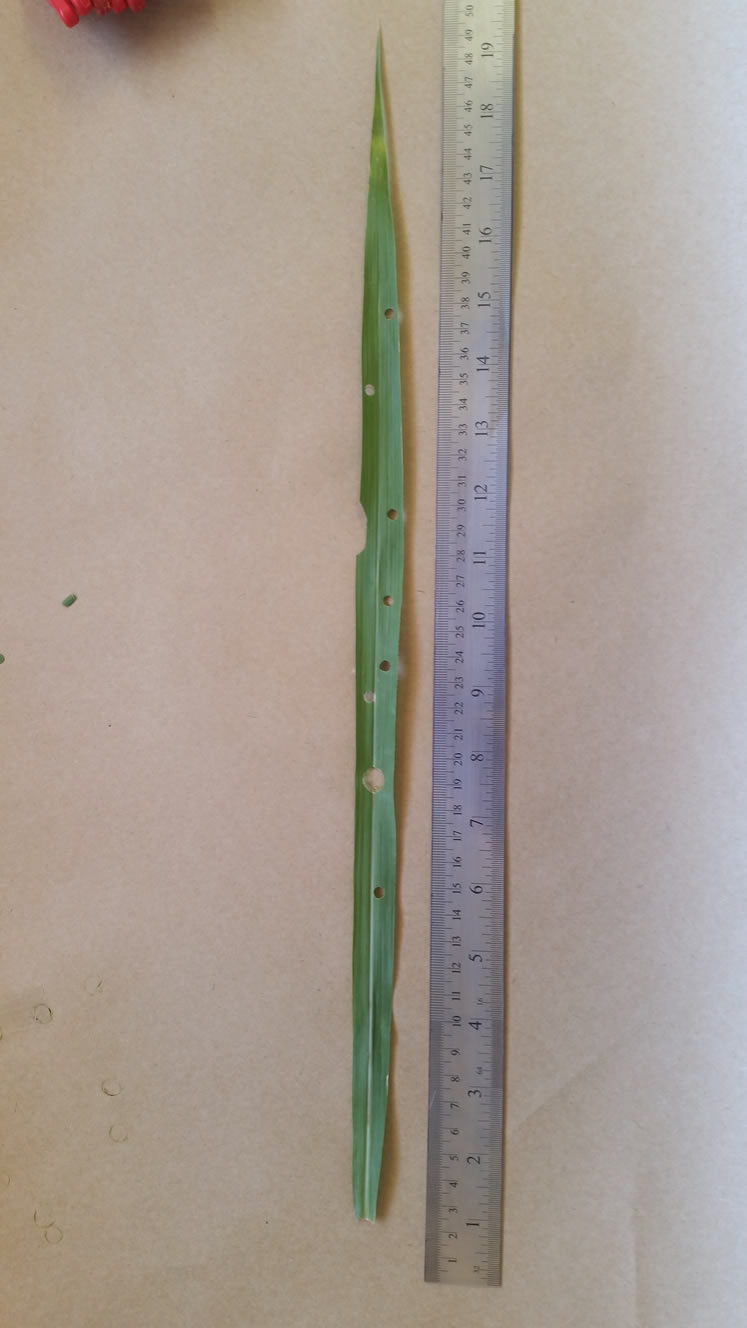}}
			\caption{\label{fig:narrowLeaves}Experiments for narrow leaf measurements of four species of plants.}
		\end{figure*}

	\medskip\noindent\textbf{Narrow leaves measurement with BioLeaf}: In this experiment, we show how our mobile application works on the measurement of leaves that are characteristically narrow. Again, with the aid of a color-contrasting background, Figure \ref{fig:narrowLeaves} illustrates the process under different illumination conditions. We experiment on four species of narrow-leaf plants, including \textit{Brachiara brizantha}, sugar cane, \textit{Brachiara marandu} and \textit{Panicum maximum}. We present the leaves and their respective defoliation estimative. With this experiment, we were able to determinate the leaf limit of the BioLeaf application. Although we can measure leaves with any size, when using regular lenses, we recommend experts to consider leaves with up to 0.5 meter.
	
	\section{Comparison with LI-COR 3100}
	\label{sec:comparisonLICOR}
	In this experiment, we directly compare equipment LI-COR 3100 and our mobile application. To this end, we created three experiments: (1) artificial herbivory with regular damages, (2) artificial herbivory with irregular damages and, (3) natural herbivory made by insects.
	
\begin{figure*}[!ht]
		\centering
		\subfigure{\includegraphics[width=0.3\textwidth]{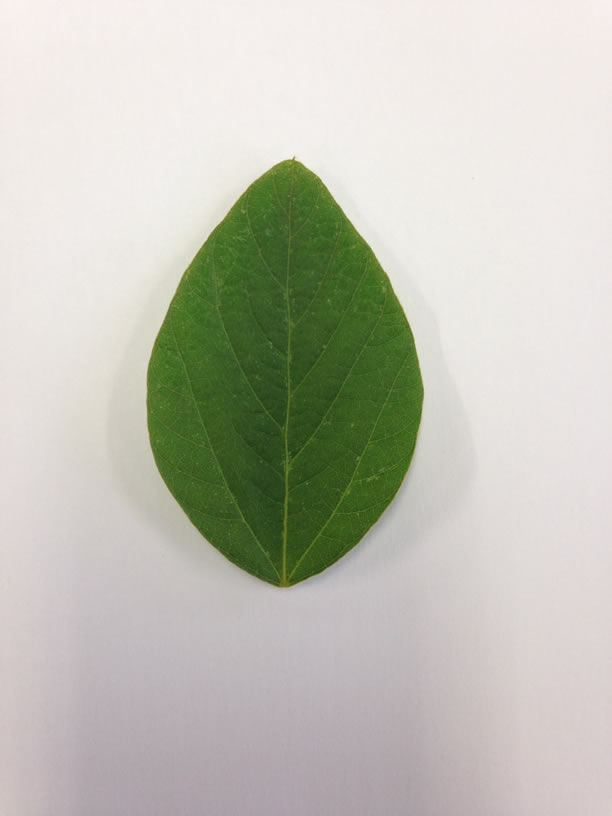}}
		\subfigure{\includegraphics[width=0.3\textwidth]{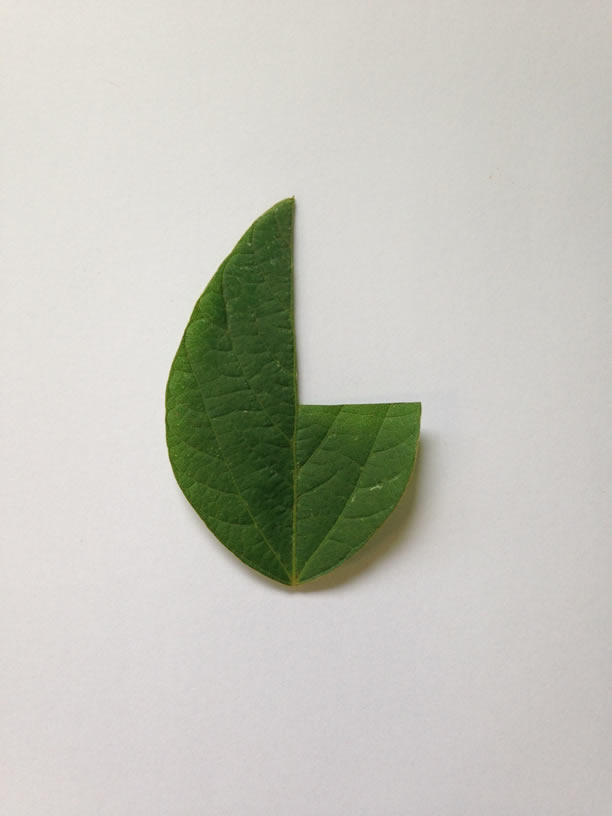}}
		\subfigure{\includegraphics[width=0.3\textwidth]{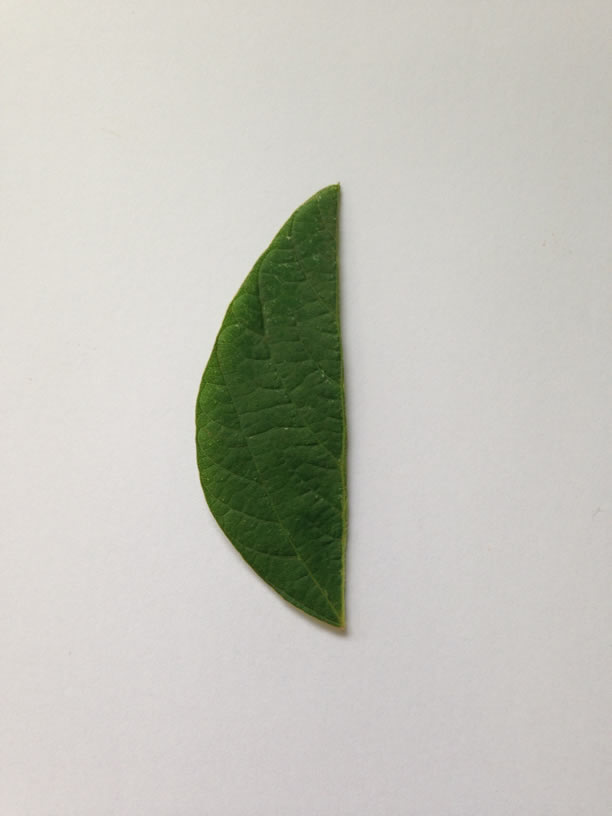}}
		\\
		\setcounter{subfigure}{0}
		\subfigure[0\%]{\includegraphics[width=0.3\textwidth]{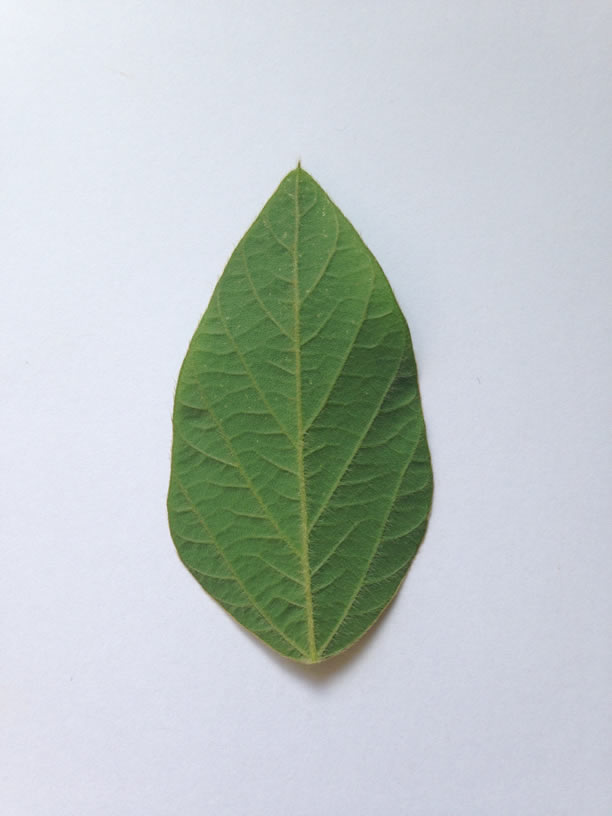}}
		\subfigure[25\%]{\includegraphics[width=0.3\textwidth]{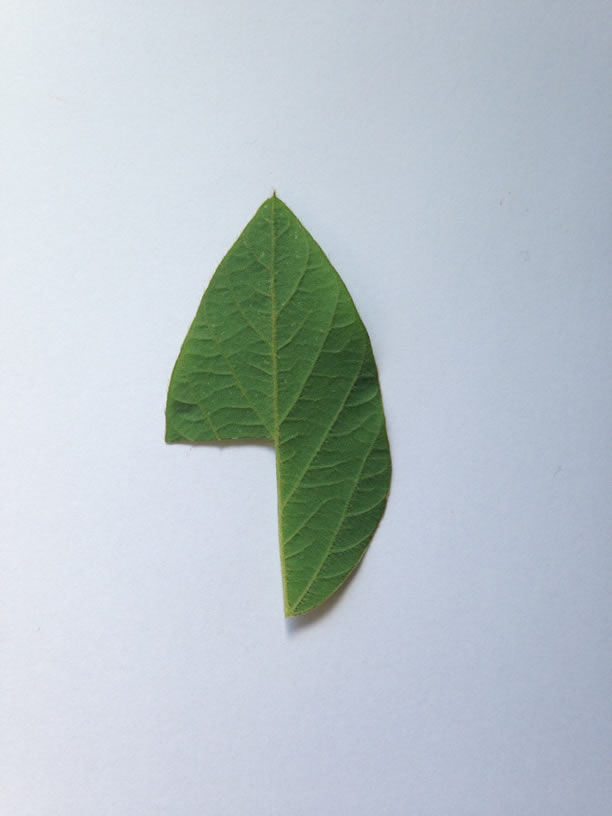}}
		\subfigure[50\%]{\includegraphics[width=0.3\textwidth]{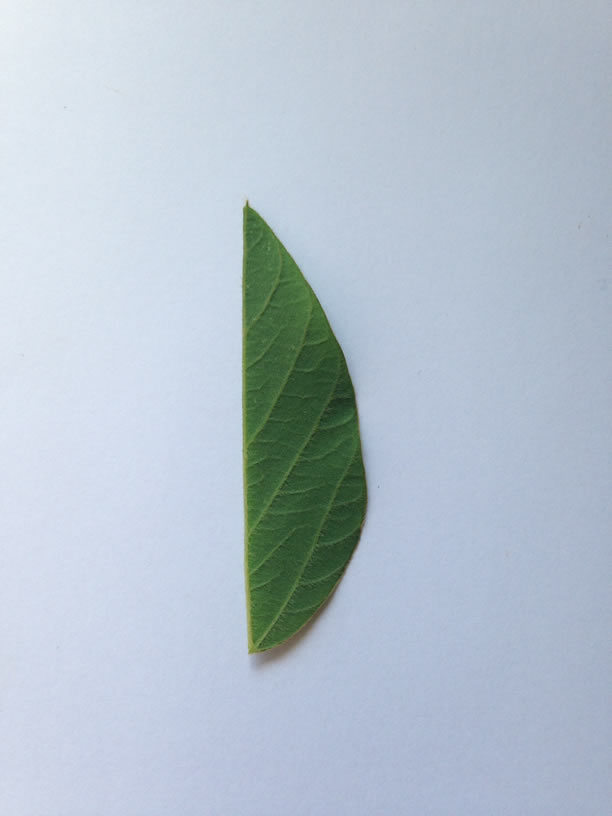}}
		\caption{\label{fig:artificialDefoliaton1}Two leaf samples with artificial herbivory and regular damages.}
	\end{figure*}

	\medskip\noindent\textbf{Artificial herbivory with regular damages}: in this case we manually created damages by using a pair of scissors, initially cutting either the base or the tip of each leaf with approximately 25\% of biomass removed; that is, a single cut with a quarter removed from the leaf. After measurements with LICOR and BioLeaf, we proceeded with a second round of measurements, now extracting approximately another 25\%, totalizing  50\% of each leaf. Figure \ref{fig:artificialDefoliaton1} illustrates this kind of damage.
	
\begin{figure*}[!ht]
		\centering
		\subfigure[\label{fig:exp1-25}$25\%$]{\includegraphics[width=0.45\textwidth]{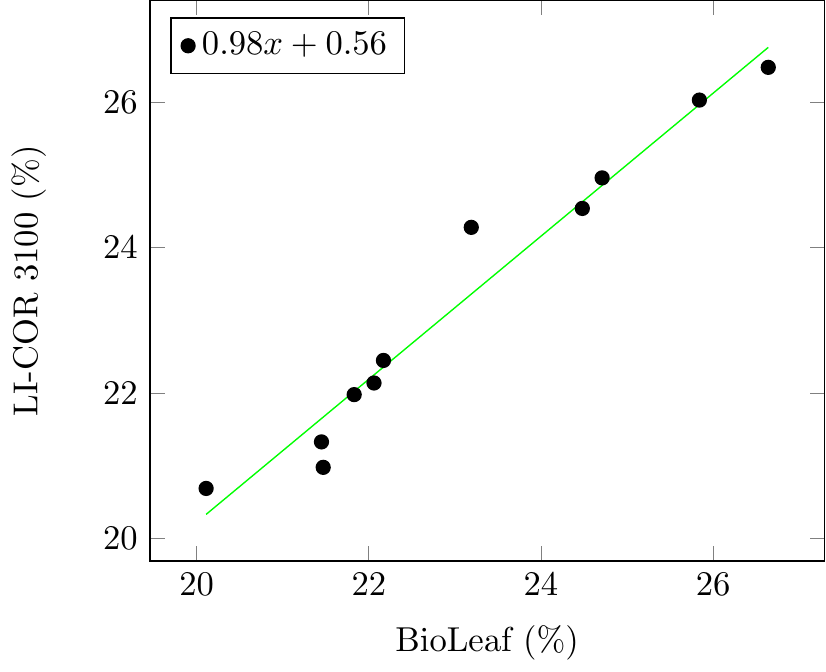}}
		\subfigure[\label{fig:exp1-50} $50\%$]{\includegraphics[width=0.45\textwidth]{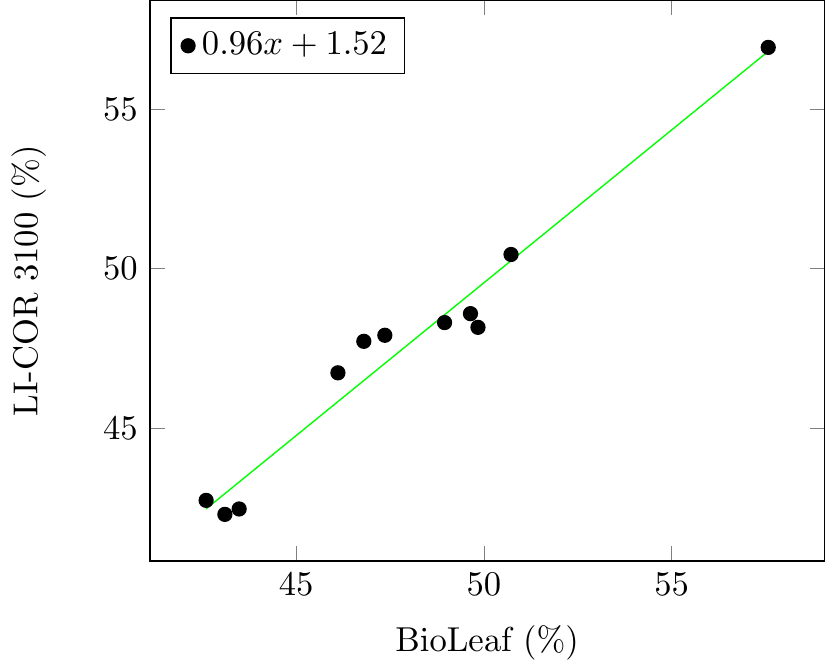}}
		\caption{Linear correlation plot demonstrating high accuracy for leaves with regular artificial damage, for $25\%$ (a) and $50\%$ (b) of damaged area.}
	\end{figure*}

	After the two series of measures, we evaluated the accuracy by analyzing the correlation \cite{gibbons1985} between the results of LI-COR 3100 and of BioLeaf (see Figures \ref{fig:exp1-25} and \ref{fig:exp1-50}). We verified a linear correlation with no significant divergence at any leaf size. The difference of the standard deviation of LI-COR and BioLeaf had a precision in the range of $\pm0.01\%$ for $25\%$ and  $\pm0.02\%$  for $50\%$.
	
\begin{figure*}[!ht]
		\centering
		\subfigure[\label{fig:exp2}Irregular damage]{\includegraphics[width=0.45\textwidth]{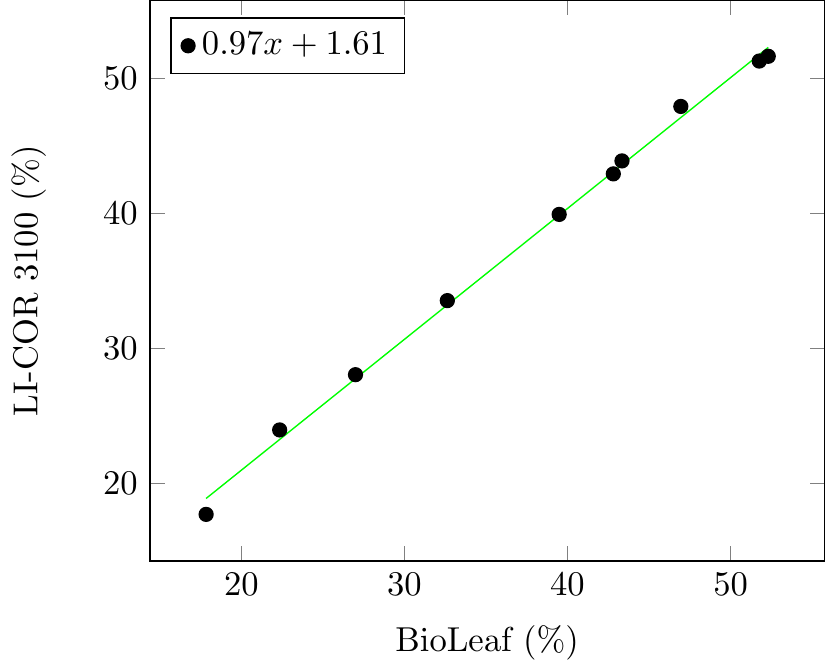}}
		\subfigure[\label{fig:exp3}Natural herbivory]{\includegraphics[width=0.45\textwidth]{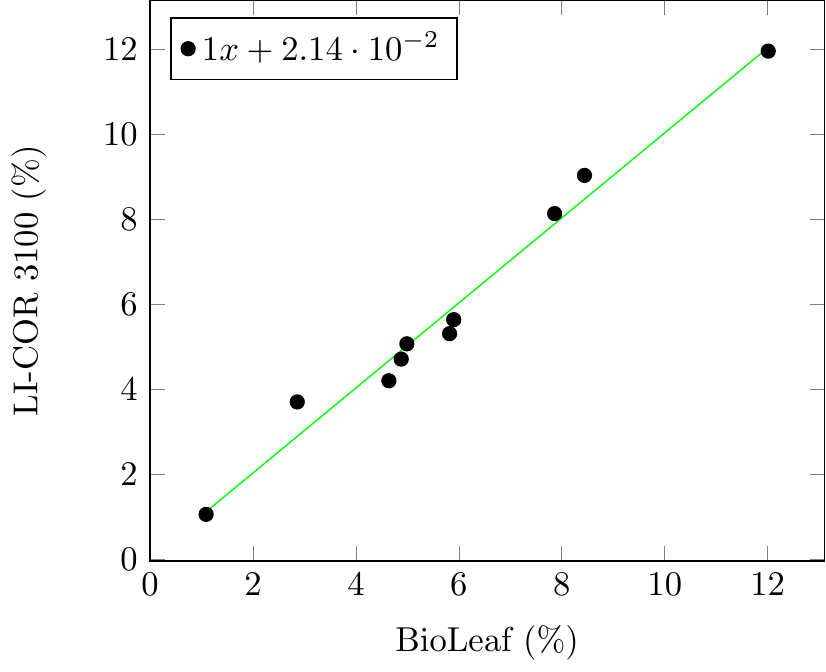}}
		\caption{Linear correlation plot demonstrating high accuracy for leaves with (a) irregular artificial damage and (b) natural herbivory caused by insects.}
	\end{figure*}

	\medskip\noindent\textbf{Artificial herbivory with irregular damages}: in this case, we manually cut either multiple, circular holes, or a single hole of a leaf blade. Additionally, in some leaves, we used scissors to produce border damages. We varied the diameters of the holes from 0.8 cm to 2.6 cm. It is important to mention that we did not choose a specific location to make the holes. The main difference to the previous experiment is the fact that we produced random internal holes together with damages in the border throughout the leaf blades. We evaluated the accuracy again using the linear correlation \cite{gibbons1985} for LI-COR 3100 versus BioLeaf (see Figure \ref{fig:exp2}). No significant divergence was observed. The difference between the standard deviations of LI-COR and BioLeaf was in the range of $\pm0.37\%$.
	
	\begin{figure*}[!ht]
		\centering
		\subfigure{\includegraphics[width=0.3\textwidth]{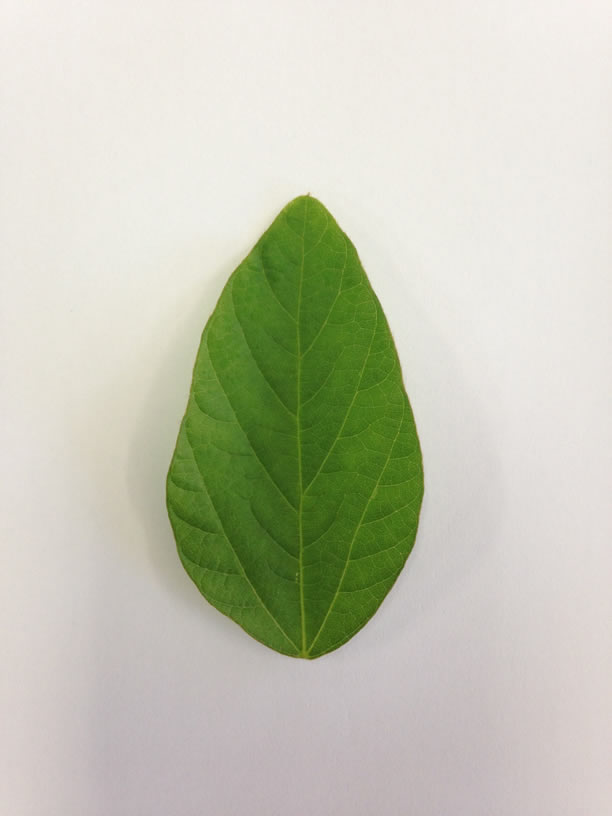}}
		\subfigure{\includegraphics[width=0.3\textwidth]{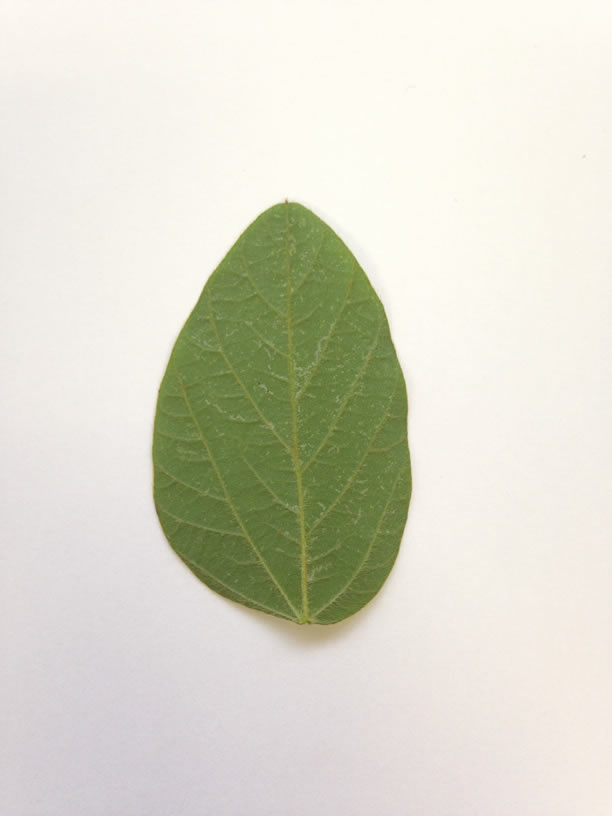}}
		\subfigure{\includegraphics[width=0.3\textwidth]{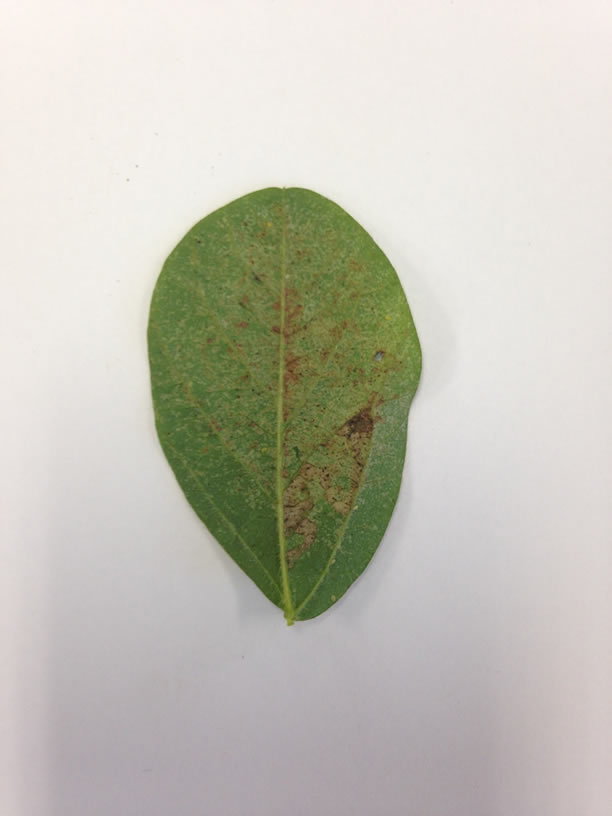}}    
		\\        
		\subfigure{\includegraphics[width=0.3\textwidth]{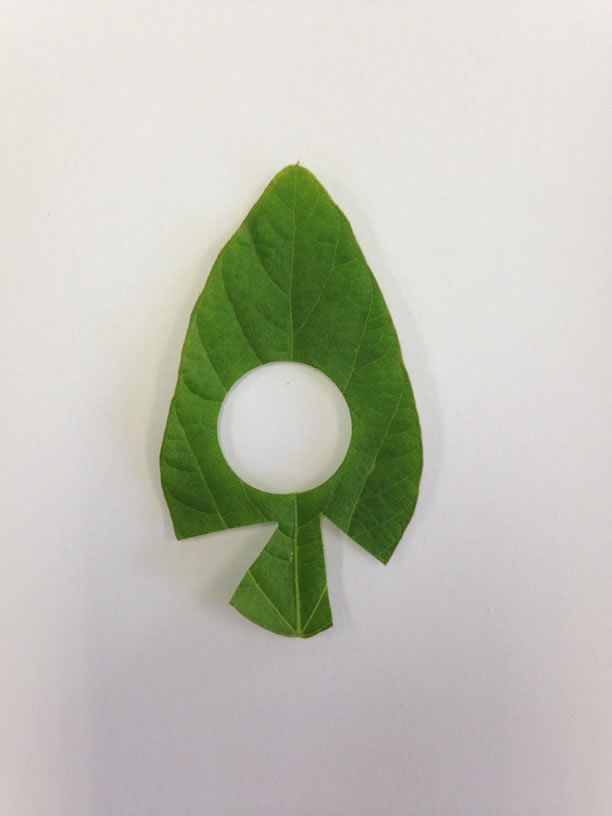}}
		\subfigure{\includegraphics[width=0.3\textwidth]{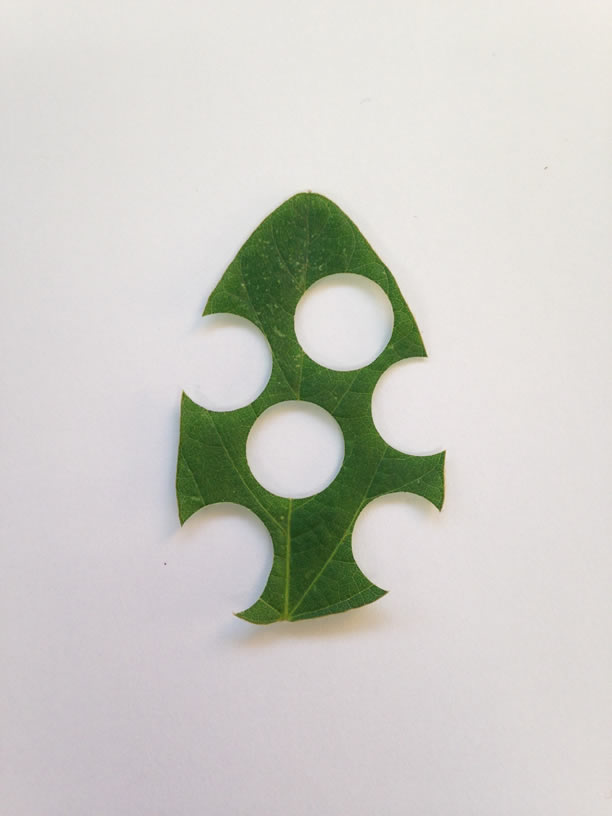}}
		\subfigure{\includegraphics[width=0.3\textwidth]{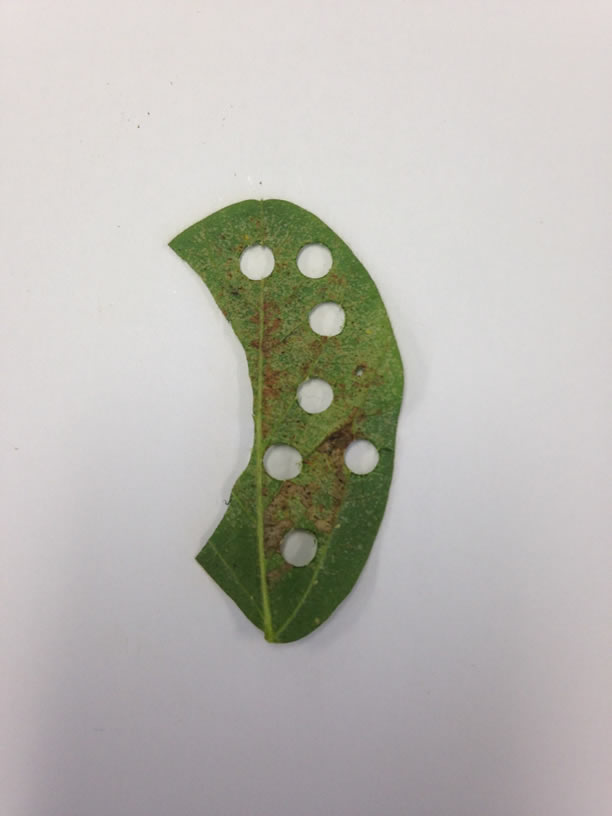}}    
		\caption{\label{fig:artificialDefoliaton2}Three leaf samples with artificial herbivory and irregular damages.}
	\end{figure*}
	
	\medskip\noindent\textbf{Natural herbivory}: in this case, the damage was done by insects only, which caused internal and, in some cases, also border attacks. The accuracy was evaluated again with the linear correlation for LI-COR 3100 versus BioLeaf (see Figure \ref{fig:exp3}). This time, the difference between the standard deviations of LI-COR and BioLeaf was in the range of $\pm0.4\%$.
	
	This set of experiments was able to demonstrate that Bioleaf is capable of substituting expensive heavy equipment that, prior to mobile processing, was the only choice to evaluate crop herbivory. It is our contention that the future of crop management will be based on mobile technology, Bioleaf, in this aspect, is in the forefront of the next technology improvements.
	
	\section{Discussion of the Results}
	\label{sec:discussao}
	
	We presented a methodology to quantify the foliar damage observed in soybean leaves answering to the large demand of such estimation in one of the most important agricultural crops in the world. We instantiated this methodology in a mobile app named Bioleaf, a new tool freely distributed online for Android systems.
	Evaluation of the BioLeaf of two cases of damages in the soybean crop \textit {Glycine max (L.) Merrill} -- with borders and when borders need to be reconstructed -- demonstrated that this application was highly accurate when compared with manual quantification. Linear correlations were significantly higher in the leaves whose borders were preserved, with correlation coefficients $R \geq 99.76$ and $P$-$value < 0.001$ (Figure \ref{fig:interna}). Correlation coefficients for borders reconstructed were slightly lower, $R \geq 99.24$ and $P$-$value < 0.001$ (Figure \ref{fig:curva}), among the six groups tested.
	
	The BioLeaf application can be used as a non-destructive tool because it does not require leaf removal from the plant, which allows repeated measurements of the same leaf. In addition, our application can handle the presence of noise that can appear in the image acquisition, such as grains of sand and even small parts of leaves. Our application is also able to reconstruct the contours of damaged leaves for multiple types of damages caused by insects.
	Therefore, the experiments showed that our tool has successfully quantified the attacked areas, regardless of the attacks being internal-only to the leaves, or concerning the borders as well. Belief can run on any Android system and it is available on GooglePlay, to date, the biggest commercial website for Android applications. The cost of our solution contrasts to that of commercial alternatives -- desktop or portable area-integrating metters -- which is above \$12,000. Mobile phones, on the other hand, are ubiquitous and accessible for popular prices.

	Besides the statistical evaluation and accuracy of BioLeaf, it is efficient in terms of processing cost, running smoothly in several mainstream mobile phones, as the Samsung S4, S6 Edge and Sony Z2 Xperia. Another potential advantage is the availability of BioLeaf on a mobile platform, which allows flexibility for on-site image collection for further analyses.
	
	It is worth saying that BioLeaf is not limited to the analysis of soybean leaves. For instance, a user can quantify the damaged areas of several  agricultural crops of similar leaf size, including cotton, bean, potato, coffee, and vegetables; as well as for monitoring attacks of different species insects, such as \textit{Helicoverpa armigera} that has lately been considered a big threat to Brazilian soybean crops.
	
	\section{Conclusions}
	\label{sec:conclusao}
	
	We introduced BioLeaf, a semi-automatic, interactive, multi-language, and portable application to estimate the herbivory of leaves. We conclude that the precision of BioLeaf was enough to allow the use of mobile phones as substitutes for expensive machinery in the task of estimating herbivory damage. The methodology was instantiated in a reliable tool (available at \url{https://play.google.com/store/apps/details?id=upvision.bioleaf}) for leaf damage measurement for use \textit{in situ} and without the removal of leaves. We tested our tool with soybean leaves, but preliminary experiments demonstrated its adequacy for use with different crops.
	
	\section*{Acknowledgements}
	This research was supported by the Dom Bosco Catholic University (UCDB) and by the Fundação de Apoio ao Desenvolvimento do Ensino, Ciência e Tecnologia do Estado de Mato Grosso do Sul, FUNDECT. Some authors also received grants from Conselho Nacional de Desenvolvimento Científico e Tecnológico, CNPQ, and from Coordenação de Aperfeiçoamento de Pessoal de Nível Superior, CAPES. The authors are thankful to Dr. Jose Valerio, at the EMBRAPA Beef Cattle - Brazilian Agricultural Research Corporation for his assistance during the execution of the experiments.
	
	\bibliographystyle{plain}

	
\end{document}